\documentclass[manuscript,screen]{acmart}

\AtBeginDocument{%
  }

\usepackage{amsmath,amsfonts}
\usepackage{textcomp}
\usepackage{url}
\usepackage{verbatim}
\usepackage{graphicx}
\usepackage{multirow}
\usepackage{makecell}
\usepackage{tabularx,booktabs}
\usepackage{subcaption}
\begin{document}

\title{Joint-Dataset Learning and Cross-Consistent Regularization for Text-to-Motion Retrieval}

\author{Nicola Messina}
\authornote{Both authors contributed equally to this research.}
\email{nicola.messina@isti.cnr.it}
\orcid{0000-0003-3011-2487}
\affiliation{%
  \institution{ISTI-CNR}
  \city{Pisa}
  \country{Italy}
}

\author{Jan Sedmidubsky}
\authornotemark[1]
\email{sedmidubsky@mail.muni.cz}
\orcid{0000-0002-7668-8521}
\affiliation{%
  \institution{Masaryk University}
  \city{Brno}
  \country{Czechia}
}

\author{Fabrizio Falchi}
\email{fabrizio.falchi@isti.cnr.it}
\orcid{0000-0001-6258-5313}
\affiliation{%
  \institution{ISTI-CNR}
  \city{Pisa}
  \country{Italy}
}

\author{Tom\'{a}\v{s} Rebok}
\email{rebok@ics.muni.cz}
\orcid{0000-0002-2331-7671}
\affiliation{%
  \institution{Masaryk University}
  \city{Brno}
  \country{Czechia}
}


\newcommand\ourarchitecture{MoT++}
\newcommand\ourloss{CCCL}
\newcommand\ourlosse{Cross-Consistent Contrastive Loss}

\begin{abstract}
Pose-estimation methods enable extracting human motion from common videos in the structured form of 3D skeleton sequences. Despite great application opportunities, effective content-based access to such spatio-temporal motion data is a challenging problem. In this paper, we focus on the recently introduced text-motion retrieval tasks, which aim to search for database motions that are the most relevant to a specified natural-language textual description (\textit{text-to-motion}) and vice-versa (\textit{motion-to-text}). 
Despite recent efforts to explore these promising avenues, a primary challenge remains the insufficient data available to train robust text-motion models effectively. To address this issue, we propose to investigate joint-dataset learning -- where we train on multiple text-motion datasets simultaneously -- together with the introduction of a \ourlosse{} function (\ourloss), which regularizes the learned text-motion common space by imposing uni-modal constraints that augment the representation ability of the trained network. 
To learn a proper motion representation, we also introduce a transformer-based motion encoder, called \ourarchitecture{}, which employs spatio-temporal attention to process sequences of skeleton data. 
We demonstrate the benefits of the proposed approaches on the widely-used KIT Motion-Language and HumanML3D datasets. We perform detailed experimentation on joint-dataset learning and cross-dataset scenarios, showing the effectiveness of each introduced module in a carefully conducted ablation study and, in turn, pointing out the limitations of state-of-the-art methods.
\end{abstract}



\maketitle


\section{Introduction}

\begin{figure}
  \centering
  \includegraphics[width=\textwidth,page=7]{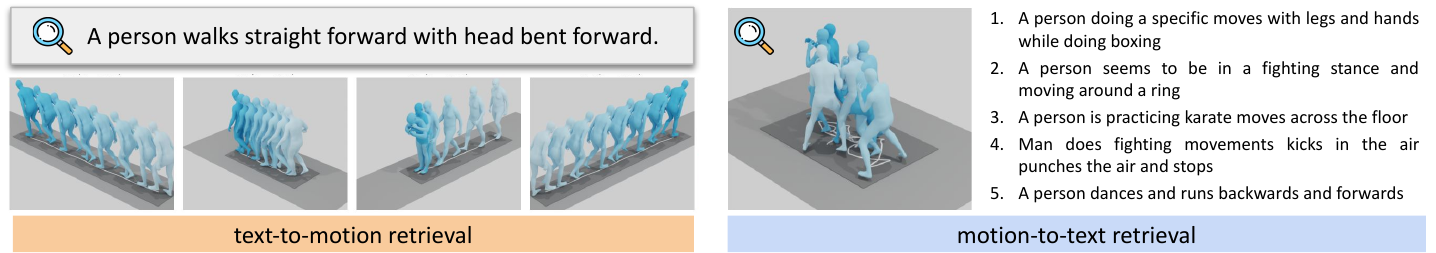}
  \caption{Formulation of the tasks: text-to-motion retrieval (left) and motion-to-text retrieval (right).}
  \label{fig:teaser}
\end{figure}

Pose estimation methods~\cite{DD22} reconstruct the virtual 3D positions of human-body keypoints from a single-camera video stream. The positions of keypoints estimated in individual frames constitute a simplified spatio-temporal representation of human motion known as a \emph{skeleton sequence}.
%
Analyzing human motion through this skeleton modality offers several advantages over the video modality~\cite{WDYD23}, including higher motion abstraction, much less sensitivity to background information, more consistent representation across different viewpoints, and much better computational efficiency~\cite{MSFR23-sigir}.
As indicated in~\cite{SEBZ21-survey}, the analysis of the skeleton representation unlocks unprecedented application potential in many domains.
For example, the skeleton data could be used in telemedicine to remotely evaluate a patient's progress in rehabilitation, in sports to assess a figure-skating performance without the emotions of human referees, in smart cities to detect potential threats from surveillance cameras like a running group of people, in virtual reality to transpose real movements into virtual environments, or in robotics to study and develop humanoid robots and human-robot interfaces~\cite{SEBZ21-survey}.
%
The ever-increasing popularity of skeleton data calls for technologies able to semantically analyze large volumes of such spatio-temporal data with respect to the focus of a target application.


%
Current research in skeleton-data analysis primarily focuses on designing deep-learning architectures for classification of actions~\cite{WLG24,ZQDDCQC23,WLGGWL23} or detection of such actions in continuous streams~\cite{GQD23,TJZLT23,PGBA19}.
%
The trained network architectures
can then serve as \emph{motion encoders} that express the motion semantics by a high-dimensional \emph{embedding} (i.e., \emph{feature vector}) extracted from the last hidden network layer. To compute the similarity between a pair of motions, the distance between their embeddings is calculated, e.g., using the cosine or Euclidean distance functions. Traditional content-based motion retrieval methods~\cite{SEBZ21-survey,JLHYJP23,BSZ21-icmr} are based on the \emph{query-by-example} paradigm, which aims at identifying the database motions that are the most similar to a user-defined query motion example. However, specifying a convenient query motion example may be problematic or even impossible since such an example might not ever exist.


These challenges inspire the development of smarter techniques for understanding and accessing 3D motion data. In this paper, we concentrate on learning a latent interaction among motion and text modalities, focusing on the challenging \emph{text-to-motion retrieval} task -- which aims at searching a database of skeleton sequences and determining those that are the most relevant to a textual query formulated in natural language description -- as well as its symmetric \emph{motion-to-text retrieval} variant, which aims at finding the text sentences in a database mostly relevant to a given query motion, as illustrated in Fig.~\ref{fig:teaser}. Specifically, text-to-motion retrieval has nice downstream applications for efficiently and effectively browsing large motion collections without relying on the query-by-example paradigm. This task was introduced in some recent works ~\cite{MSFR23-sigir,PBV23-TMR} and is still relatively underexplored. The objective of the original idea~\cite{MSFR23-sigir} is to separately encode motion and text modalities and project the obtained motion and text latent representations into the same common space. We build upon this idea~\cite{MSFR23-sigir} and incorporate some common retrieval benchmarking protocols employed in~\cite{PBV23-TMR}. We also introduce an improved transformer-based motion encoder called \ourarchitecture, as the extension of MoT proposed in~\cite{MSFR23-sigir}. One of the key contributions of this work is to solve a principled problem arising in this domain, which is the lack of sufficient data to obtain good motion-to-text and text-to-motion retrieval generalization. In order to achieve this goal, we progress towards scenarios that employ the two key datasets in this domain as training data in a \emph{joint-dataset learning} setup. Furthermore, we enhance the training process with a new loss function, which we call \ourlosse\ (\ourloss), that regularizes the learned embedding space by imposing some cross-consistency constraints among the scores computed within and across modalities to improve generalization.

%
\section{Related Work}

For the adopted text-to-motion retrieval task, we mainly need to: (1) encode both text and skeleton data modalities into compact and content-preserving latent representations, (2) learn the common multi-modal space for both the modalities, and (3) manage the motion representations so that the most relevant ones are efficiently obtained for a given text representation.
In this section, we survey relevant existing methodologies: (A) text encoder methods, (B) motion encoder methods, and (C) cross-modal techniques processing both the motion and text modalities.

\subsection{Text Encoders}

Recent advancements in neural language models strongly help solve various tasks in natural language processing. The mainstream is to pre-train Transformer models over large-scale corpora. The resulting models differ in size, ranging from smaller (e.g., BERT~\cite{kenton2019bert} with 110 million parameters) to large language models (e.g., GPT-3~\cite{brown20-gpt3} with 175 billion parameters).
Since the domain of human motion is particularly limited, it is sufficient to employ small-scale models.
%
In~\cite{ghosh2021synthesis}, a BERT-based language model is employed within the motion synthesis task conditioned on a natural language prompt. This model stacks together a BERT pre-trained module and an LSTM model composed of two layers for aggregating the BERT output tokens, producing the final text embedding.
In the text-to-motion retrieval task~\cite{MSFR23-sigir}, the final hidden state of the LSTM model is considered as the final sentence representation.
%
CLIP~\cite{RKHRGASAMCKS21-CLIP} is the vision-language model trained in a contrastive manner for projecting images and natural language descriptions in the same common space.
The textual encoder of this model is composed of a transformer encoder~\cite{vaswani2017attention} with modifications introduced in ~\cite{radford2019language}, and employs lower-cased byte pair encoding (BPE) representation of the text. 
This encoder is also utilized in text-to-motion retrieval~\cite{MSFR23-sigir} by stacking an affine projection to the CLIP representation.
%
In~\cite{PBV23-TMR}, the authors employ a simple transformer-based encoder inspired by the VAE encoder in~\cite{PBV21}. This encoder, called \textit{ACTORStyleEncoder}, takes input text sequences of arbitrary length concatenated to two learnable \texttt{[class]} tokens. In output, it employs these two special tokens to estimate the mean and variance vectors of a multinomial Gaussian distribution, used by the decoder to reconstruct the corresponding motion.
Despite not being pre-trained on large textual corpora, ACTORStyleEncoder seems better able to capture motion-related textual descriptions, outperforming CLIP in this domain. For this reason, we also employ such an encoder to obtain a suitable textual representation.
%

\subsection{Motion Encoders}

Motion encoders have traditionally learned latent motion representations in a supervised way for the classification task, mainly based on transformers~\cite{WLG24,MSFR23-sigir,AKC21,CCCWZL21,CCZL20}, convolutional~\cite{LWFP21}, recurrent~\cite{SLXZL18}, or graph-convolutional~\cite{ZQDDCQC23,DWCL22} networks, or their combinations (e.g., transformer and 3D-CNN~\cite{STS23}). The current trend is to rely on self-supervised methods~\cite{WLGGWL23, SLDQGYWW23, SCLLHQ23}, as they can learn motion semantics without knowledge of labels using reconstruction-based or contrastive-based learning.
The reconstruction-based approach~\cite{ZZCHZZLS23,SCA23-ecir} applies the encoder-decoder principle to reconstruct the original skeleton data of an input motion and uses the learned intermediate feature as the latent representation. The contrastive-learning approach~\cite{YLG22,LSYL20} aims at learning a meaningful metric that sufficiently reflects semantic similarity to discriminate motions belonging to different classes in the validation step.

To increase the descriptive power of the latent representations, the learning process can integrate other skeleton modalities that are extracted from 3D coordinates and provide complementary information to the original joint modality, as recently surveyed in~\cite{SCIA24}. For example, early-fusion strategy in~\cite{SLDQGYWW23} is applied to jointly encode the joint, bone, and motion modalities in a single-stream manner. In~\cite{CCLRL24}, the late-fusion strategy is applied to skeleton modalities of joints and bones and also to the motion-map modality extracted from RGB frames cropped around the detected skeletons. In~\cite{LWPNK23}, RGB and depth streams are encoded using a recurrent neural network with specialized multi-modal contextualization units. In~\cite{GQD23}, a fusion module of skeleton and RGB modalities is proposed to enable the two features to guide each other.

We primarily focus on two recent motion encoders proposed in~\cite{PBV23-TMR,MSFR23-sigir}.
In~\cite{PBV23-TMR}, the authors employ the same ACTORStyleEncoder not only for text encoding but also for motion encoding, as originally proposed in~\cite{PBV21}. In the case of the motion modality, input tokens representing words in a phrase are replaced by skeleton poses across different timesteps. The motion encoder in~\cite{MSFR23-sigir} introduces a spatio-temporal transformer, called MoT, which is principally similar to ViViT for video encoding~\cite{ADHSLS21}. MoT processes both spatial and temporal features using the attention mechanism. In this work, we improve MoT by employing a different kind of spatio-temporal attention and a methodology to aggregate skeleton joints without losing feet and root information.


\subsection{Mutual Processing of Motion and Text Modalities}

The current trend in multimedia processing is to learn a common multi-modal space for the visual and textual modalities~\cite{RKHRGASAMCKS21-CLIP,messina2022transformer,fang2021clip2video,shvetsova2022everything} so that similar images or videos can be described and searched with textual descriptions. This enables the use of open vocabularies or complex textual queries to search for relevant images/videos.
Inspired by such powerful and versatile text-vision models, new works~\cite{KKPJZKK24,ZZCHZZLS23,PBV22,GZZWJLC22-HumanML3D,TRGSCB22,ZCPHGYL22,PBV21} have started to emerge also in the \emph{text-motion} domain. These works focus on \emph{motion synthesis}, i.e., generating skeleton avatars from a textual description. In contrast to video data, the skeleton modality is anonymized and avoids learning many common biases present in video datasets. The principal idea of text-motion methods is the same as the text-vision methods -- to align text and motion embeddings into the common space. This can be further enhanced by predicting a proper motion length from text~\cite{GZZWJLC22-HumanML3D} or by training text-to-motion and motion-to-text tasks jointly~\cite{GZWC22}. The recent methods~\cite{TRGSCB22,ZCPHGYL22} employ the diffusion principle to gradually add noise to a sample from the data distribution and learn the reverse process of denoising the sample by a backbone generative neural network.

Besides motion synthesis, text-motion processing has been used for improving skeleton-based \emph{classification} by text-to-motion matching~\cite{kim2022learning} or by additionally generated textual descriptions from training actions~\cite{WLG24}. Employing the text-motion modalities for retrieval purposes has not yet been studied much. There are two recent fundamental papers~\cite{MSFR23-sigir} (SIGIR 2023) and~\cite{PBV23-TMR} (ICCV 2023) that independently tackle the text-motion retrieval tasks.
Both approaches are essentially very similar as they learn a common cross-modal embedding space for both the text-motion training pairs with the InfoNCE loss function. While the former approach~\cite{MSFR23-sigir} proposes a Motion Transformer (MoT) to encode motions, the latter approach~\cite{PBV23-TMR} adopts ACTORStyleEncoder
trained for motion synthesis using contrastive learning with careful selection of negatives by filtering out the \textit{wrong} negatives -- which are the negative samples too similar to the positive one. In the retrieval phase, the embedding of a user text query is extracted and compared to the embeddings of database motions to determine the $K$ nearest-neighbor motions. Both the approaches achieve competitive retrieval results on the KIT Motion Language (KITML)~\cite{Plappert2016-KITdataset} and HumanML3D~\cite{GZZWJLC22-HumanML3D} benchmark datasets, that provide both text and skeleton-data modalities. A very similar subsequent work~\cite{YLWDLL23} aims to learn the common space properties more effectively by carefully selecting hard negative samples during triplet-loss training.

\subsection*{Contributions of this Paper}

The two mainstream text-to-motion retrieval approaches~\cite{MSFR23-sigir,PBV23-TMR} learn the text-motion representations independently for each dataset, either KITML or HumanML3D. Although the \emph{cross-dataset} (i.e., training dataset differs from validation dataset) or \emph{joint-dataset} (i.e., training and validation datasets come from the fusion of multiple diverse datasets) learning approaches are principally known~\cite{TLYZLZ22}, they have not been tackled in the text-motion domain. Studying the generalization abilities of retrieval methods in low-data regimes is of critical importance for achieving good and reliable models.

In light of the above, we contribute to the field
in two principal ways: (1) by proposing \ourarchitecture, an improvement of MoT~\cite{MSFR23-sigir}, which includes a more effective spatio-temporal attention mechanism and a more nuanced joint aggregation schema, and (2) by tackling the lack of motion-text data by studying \emph{joint-dataset} and \emph{cross-dataset} generalization,
together with a new loss function that regularizes the learned common space by also enforcing uni-modal score constraints.
Our contributions can be summarized as follows:
\begin{itemize}
    \item We introduce motion encoder \ourarchitecture{} that improves on MoT~\cite{MSFR23-sigir} by integrating effective spatio-temporal attention schemas and preserving information about feet and root joints better.
    \item We explore \emph{cross-dataset} evaluation and \emph{joint-dataset} learning to understand and mitigate generalization issues due to the lack of sufficiently diverse data.
    \item We propose a new loss function, called \ourlosse\ (\ourloss), that better constrains the common space by augmenting the InfoNCE objective with uni-modal loss terms to mitigate the data scarcity issue more effectively.
    \item We perform an extensive experimental evaluation on single-dataset, cross-dataset, and joint-dataset learning scenarios. We establish new baselines, especially for cross-dataset and joint-dataset learning.
    \item We experimentally compare our approach with state-of-the-art methods using benchmarks defined for the KITML and HumanML3D datasets.
\end{itemize}
%


\begin{figure}[t]
  \centering
  \includegraphics[width=\linewidth,page=8]{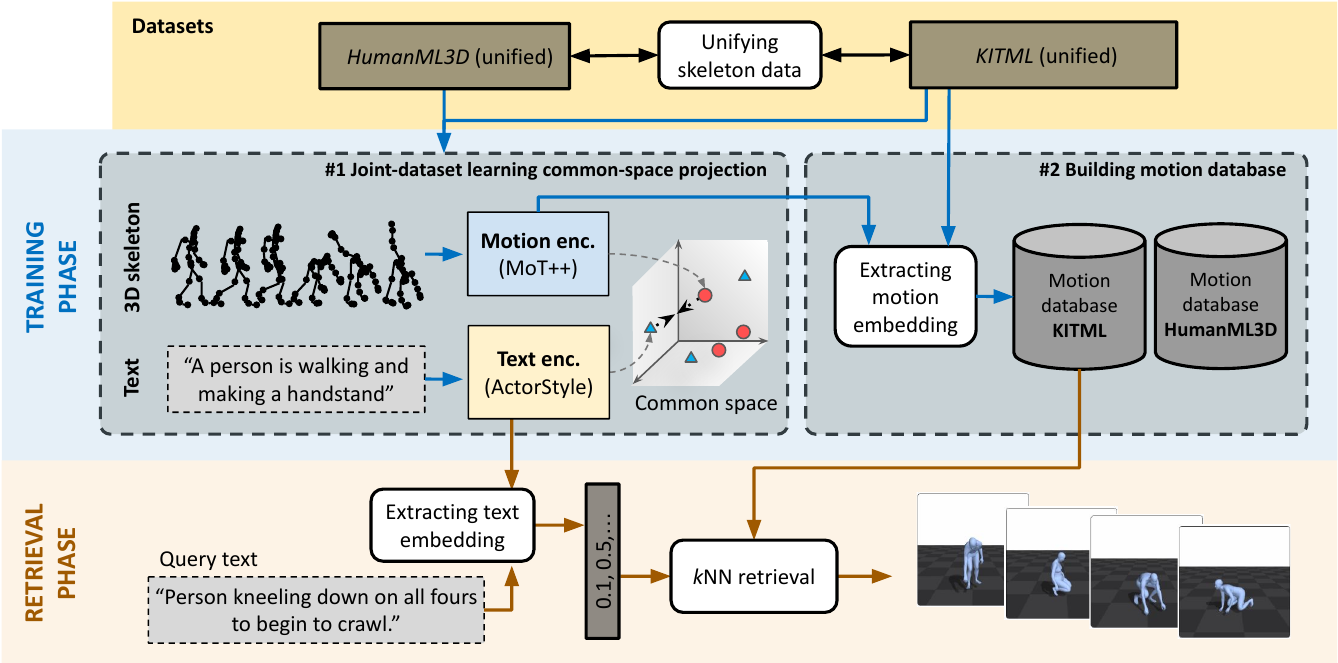}
  \caption{Schematic illustration of the whole architecture. In the training phase, (1) joint-dataset learning is applied to unified HumanML3D and KITML datasets to learn the common space of both the text and motion modalities and (2) the trained motion encoder is then used to extract motions' embeddings that are stored in a database (exemplified on the KITML dataset). In the retrieval phase, the embedding of a given text query is extracted and compared against the embeddings of motions in the KITML database to retrieve the $K$ most relevant motions.}
  \label{fig:schema}
\end{figure}

\section{Text-to-Motion and Motion-to-Text Learning Pipeline}

Our text-to-motion retrieval approach is principally similar to that of state-of-the-art works in this field~\cite{PBV23-TMR,MSFR23-sigir}. In particular, it is a two-stream learning pipeline, where text and motion features are first extracted through ad-hoc encoders and then projected into the same common space. We especially focus on joint-dataset learning to achieve better generalizability of the learned features, as schematically illustrated in Fig.~\ref{fig:schema}. In this section, we sketch the components of the whole architecture and primarily focus on our contributions with respect to both state-of-the-art works, especially the new motion encoder, loss function, and joint-dataset learning.

\subsection{Problem Definition}

We are given a database of $N$ text-motion pairs, defined as $\mathcal{S} = \{T_i, M_i\}_{i=1}^N$, where $M_i$ is the $i$-th motion and $T_i$ its corresponding textual description specified in natural language, like \textit{"A person kneeling down on all fours to begin to crawl"}. With such data organization, two symmetric tasks are defined. In \emph{text-to-motion} retrieval, a $T_i\in\mathcal{S}$ is used as a query to retrieve the $K$ most relevant motions $\{M_j\}_{j=1}^K$ from $\mathcal{S}$ based on how closely they align with the semantics of text. Conversely, in the symmetric \emph{motion-to-text} retrieval task, a query motion $M_i\in\mathcal{S}$ is used to search for the most relevant textual descriptions $\{T_j\}_{j=1}^K$ from $\mathcal{S}$.

These retrieval tasks involve extracting the query feature using a proper motion or text encoder and then comparing it against the pre-extracted features of all database motions based on a pre-defined similarity function, such as the cosine similarity. This formulation, employed in many cross-modal retrieval scenarios \cite{luo2022clip4clip,fang2021clip2video,messina2022aladin,RKHRGASAMCKS21-CLIP,jia2021scaling,MSFR23-sigir,PBV23-TMR} is very effective and efficient. In fact, if the motion database is very large, any vector-based indexing technique,
such as FAISS\footnote{https://ai.meta.com/tools/faiss/}, can be principally adopted to speed up the motion retrieval process.

In this setup, we construct a probabilistic common embedding space as introduced in \cite{PBV23-TMR}. Specifically, we extract the following quantities from each $T_i$ and $M_i$ in $\mathcal{S}$:
\begin{align}
    \mathbf{m}_i^\mu, \mathbf{m}_i^{\sigma^2} &= \mathcal{E}_m(M_i) \\
    \mathbf{t}_i^\mu, \mathbf{t}_i^{\sigma^2} &= \mathcal{E}_t(T_i)
\end{align}
where $\mathcal{E}_m$ and $\mathcal{E}_t$ are two deep neural networks having their own trainable parameters that digest motions $M_i$ and texts $T_i$, respectively. These networks output two vectors each. Specifically, $\mathbf{m}_i^\mu, \mathbf{m}_i^{\sigma^2}$ represent the mean and variance of a Gaussian distribution living in the common space and representing motion $M_i$, while $\mathbf{t}_i^\mu, \mathbf{t}_i^{\sigma^2}$ plays the same role for text $T_i$.
This formulation is driven by the underlying variational autoencoder (VAE) \cite{kingma2019introduction} core structure of the motion pipeline, which tries to reconstruct the original motion  $\tilde{M}_i$ through a decoder $\mathcal{D}_m$ in the following way:
\begin{align}
    \tilde{M}_i = \mathcal{D}_m(\mathbf{z}) \qquad \text{where} \; \mathbf{z} \sim \mathcal{N}(\mathbf{m}_i^\mu, \mathbf{m}_i^{\sigma^2}).
\end{align}
The losses employed to constrain the latent space in the VAE include the standard KL divergence loss -- that forces both text and motion features to share similar distributions and force them to the unit normal distribution -- and a motion reconstruction loss employed on the decoder's head -- formalized by a per-joint regression objective $\mathcal{L}_\text{rec} = \sum_i ||M_i - \tilde{M}_i||_1$ -- which in this setup plays only a regularization role by stabilizing the whole training process. In fact, as our downstream task is \textit{retrieval} and not \textit{generation}, we are not interested in the outcome of the motion generation part of the network, although it has been shown to produce nice regularization effects. 
Instead, the core objective for text-to-motion retrieval and motion-to-text retrieval is the creation of a common space where motions and corresponding descriptions have high similarity, while motions and uncorrelated descriptions have low similarity. This contrastive objective is enforced by a loss $\mathcal{L}_\text{retrieval} = \text{contrastive}(\{\mathbf{m}_i^\mu, \mathbf{t}_j^\mu\}_{i,j=1}^N)$ which, in our proposal, will be detailed in Section~\ref{sec:losses}.
Notice that, as in \cite{PBV23-TMR}, only mean vectors are used as features for the motion and text in the common space instead of their Gaussian distributions used to reconstruct motions. For this reason, from this point onward, we simply refer to the $i$-th text and motion features without the $\mu$ notation, as $\mathbf{m}_i$ and $\mathbf{t}_i$.

\subsection{\ourarchitecture{} Motion Encoder}
One of the key contributions of this paper is the introduction of the novel transformer-based motion encoder \ourarchitecture, which extends its predecessor MoT presented in \cite{MSFR23-sigir}. The architecture of \ourarchitecture\ is built on top of the successful transformer-based video processing network ViViT~\cite{ADHSLS21}, which implements efficient spatio-temporal attention mechanisms, namely \textit{factorized encoder} and \textit{factorized self-attention} to factorize attention computation in either the spatial and temporal dimensions, drastically reducing the needs of memory and computational resources. In the following two subsections, we present the architecture behind \ourarchitecture\ and some details behind the processing of the skeleton features used to feed it.

\subsubsection{Architecture}
\ourarchitecture\ takes as input a raw motion sequence $\mathbf{\bar{x}} \in \mathbb{R}^{T\times J\times D}$, where $T$ is the number of frames of the motion, $J$ is the number of skeleton joints, and $D$ is the dimensionality of each joint.

First, the input is preprocessed by a function $\mathcal{J}(\mathbf{\bar{x}})$ which possibly aggregates joints together, reducing $J$ to $J'$ and increasing their dimensionality from $D$ to $D'$ to better fit the transformer pipeline. We can consider each of the $J'$ new elements as a group of skeleton joints, each represented by a new $D'$-dimensional feature. Having a new sequence of $J' < J$ joints serves two purposes: (i) it is beneficial from a computational point of view, as the spatial sequence in input to the transformer is shorter, and (ii) we found that this solution helps in reducing overfitting and increasing generalization of the whole pipeline. As the $\mathcal{J}$ function, we employ a set of independent MLPs, which separately aggregate joints from seven different parts of the skeleton, following an idea similar to \cite{ghosh2021synthesis} and further developed in our previous MoT framework \cite{MSFR23-sigir}. Specifically, the first five groups are obtained by aggregating joints from the five different parts of the human body. We then include two separate MLPs for independently processing the root bone and feet floor contact state. At the end of this step, we obtain the pre-processed motion $\mathbf{x} = \mathcal{J}(\mathbf{\bar{x}}) \in \mathbb{R}^{T\times J'\times D'}$.

The resulting sequence $\mathbf{x}$ is flattened into $\mathbb{R}^{TJ'\times D'}$ and appended to two CLS tokens $\text{CLS}_\mu$ and $\text{CLS}_{\sigma^2}$ used to follow the same encoder output interface employed by the TMR framework~\cite{PBV23-TMR}, which estimates mean and variance of the latent space.
We therefore obtain a pre-processed sequence $\mathbf{x}_s = [\text{CLS}_\mu, \text{CLS}_{\sigma^2}, \mathbf{x}] \in \mathbb{R}^{(2+TJ')\times D'}$ ready to be fed into the transformer layers.

\begin{figure}[t]
  \centering
  \includegraphics[width=\linewidth,page=1]{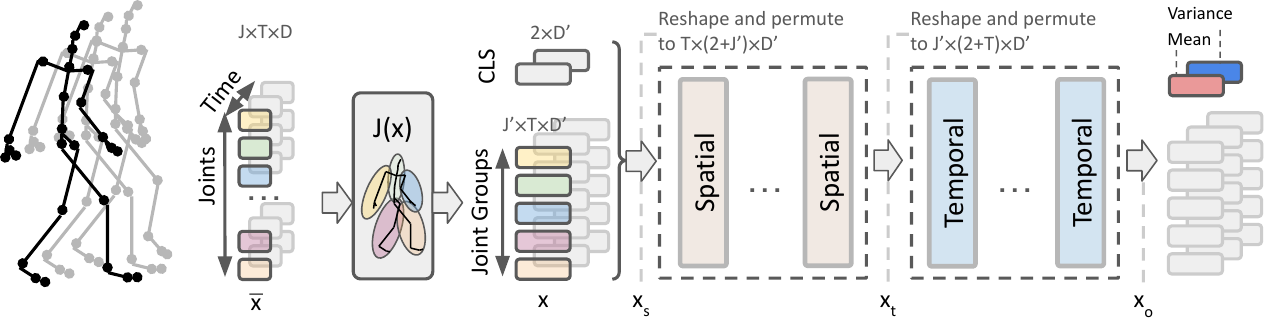}
  \caption{\ourarchitecture\ architecture. The input spatio-temporal skeleton sequence $\bar{\mathbf{x}}$ is processed by the $\mathcal{J}$ function to spatially group the skeleton joints and therefore reduce the spatial sequence length while increasing the dimensionality $D$ to $D'$. The resulting sequence $\mathbf{x}$, concatenated to two special tokens, is then processed by a spatio-temporal transformer -- in this case configured in a \textit{factorized self-attention} setup. The two CLS tokens in output are employed as $\mathbf{m}^\mu$ and $\mathbf{m}^{\sigma^2}$.}
  \label{fig:motpp_architecture}
\end{figure}

This input is processed through $N$ transformer layers, denoted by $\{\mathcal{T}_i\}_{i=1}^N$. Similarly to ViViT, each transformer layer processes either spatial information or temporal information, avoiding jointly processing space and time dimensions, which would be computationally unfeasible. For this reason, $N$ is always even so that we always have the same number of temporal and spatial layers. 

In the so-called \textit{factorized encoder} case, the first $N/2$ layers compute attention over the joint dimension to share information across skeleton joints at each timestep independently. From a practical perspective, this is obtained by simply reshaping $\mathbf{x}_s \in \mathbb{R}^{(TJ'+2)\times D'}$ into $\mathbf{x}_s^r \in \mathbb{R}^{T\times (2+J')\times D'}$, and then computing $\mathbf{x}_t = \{\mathcal{T}_i\}_{i=1}^{N/2}(\mathbf{x}_s^r)$, with the $T$ dimension playing the role of the batch size. The second $N/2$ stack of layers performs the same operation but on the time dimension for each joint. This is obtained by first permuting and reshaping $\mathbf{x}_t$ into $\mathbf{x}_t^r \in \mathbb{R}^{J'\times (2+T)\times D'}$ before seeding it into these layers, to obtain $\mathbf{x}_o = \{\mathcal{T}_i\}_{i=N/2+1}^{N}(\mathbf{x}_t^r)$, where this time $J'$ plays the role of the batch size.

The so-called \textit{factorized self-attention}, employed in MoT~\cite{MSFR23-sigir}, is very similar to the factorized encoder case, except that spatial and temporal transformer layers are interleaved: $\{\mathcal{T}_{2i}\}_{i=1}^{N/2}$ layers compute attention over time, while $\{\mathcal{T}_{2i-1}\}_{i=1}^{N/2}$ layers work out attention over the joints dimension.

From early experimentation, we found that \textit{factorized encoder} works better than \textit{factorized self-attention} on the motion domain. For this reason, we instantiate \ourarchitecture\ by employing the \textit{factorized encoder}.

The output of \ourarchitecture\ is composed of the first two tokens of the output sequence $\mathbf{x}_o$, i.e., the content of the two prepended CLS tokens that are devoted to producing the mean and variance over the latent space. Fig.~\ref{fig:motpp_architecture} shows the overall architecture of the proposed \ourarchitecture\ motion encoder.

\paragraph{Motion Features}
One of the key advantages of \ourarchitecture\ over other proposed encoders is that it employs a well-structured spatial sequence of joint tokens instead of a single flattened feature vector representing the whole skeleton. In fact, by design, \ourarchitecture\ requires a two-dimensional sequence of tokens (a spatial and a temporal one), which does not allow for a single flattened feature vector encoding the full skeleton.
In order to achieve this, the most intuitive solution is to break down the flattened representation employed in previous works \cite{PBV23-TMR,PBV22,TRGSCB22} to reconstruct a sequence of features, one for each skeleton joint. Specifically, we employ the motion vector as preprocessed by \cite{PBV23-TMR} and extract from it the $D$-dimensional feature vector for each of the $J$ different skeleton joints. Specifically, we represent each joint with $D=12$ different features, consisting of (i) a 3D sub-vector for encoding the rotation invariant forward kinematic (\textit{rifke}) motion information, encoding the spatial position of each joint without root rotation applied; (ii) a 6D sub-vector encoding the rotation information of each joint in the 6D-continuous representation format \cite{zhou2019continuity}; (iii) a 3D sub-vector encoding the spatial velocities of each joint.
This is performed for all the non-root joints of the SMPL skeleton \cite{loper2023smpl}, which corresponds to 21 joints. 
In addition to these, we also collect a root joint token -- composed of rotation velocity along the $y$-axis, the linear velocity on the $xy$ plane, and the root height -- and a feet token -- which is a virtual token carrying foot contact information, for a total of $J=23$ skeleton joints.


\subsection{\ourlosse\ (\ourloss)}
\label{sec:losses}
Although the motion encoder is a core ingredient for achieving a clever comprehension of the underlying motion semantics, the role of data and of the optimization process are also of key importance. In particular, we noticed that, due to scarcity of data, the networks are prone to overfitting if clever constraints on the learned common space. To mitigate these issues, in this work, we propose to employ more than one dataset for training so to increase the overall available number of text-motion samples -- which is anyway very small compared, for example, to the amount of data available for training image-text matching methods \cite{messina2021fine,messina2022aladin,RKHRGASAMCKS21-CLIP,zhang2021vinvl} -- 
and introducing a novel loss that highly regularizes the produced common space to deal with limited data availability. 

In this section, we focus on the introduction of this novel loss, which we call \ourlosse\ (\ourloss). It is designed to impose additional constraints with respect to the standard multi-modal contrastive learning used in previous works to better constrain the training process and, in turn, achieve higher generalization.

Specifically, given a text feature $\mathbf{t}_i$ and the corresponding motion feature $\mathbf{m}_i$, the proposed \ourloss\ loss enforces the following objectives.
\begin{itemize}
    \item A cross-modal contrastive objective, which enforces $\mathbf{t}_i$ to have a higher cosine similarity to $\mathbf{m}_i$ with respect to $\mathbf{m}_j$ ($i\neq j$) for text-to-motion retrieval and vice-versa for motion-to-text retrieval.
    \item A uni-modal similarity objective for textual descriptions, which enforces semantically similar/dissimilar texts $\mathbf{t}_i$ and $\mathbf{t}_j$ ($i\neq j$) to have high/low cosine similarity.
    \item A uni-modal similarity objective for motion features, which enforces similar/dissimilar motions $\mathbf{m}_i$ and $\mathbf{m}_j$ ($i\neq j$) to have high/low cosine similarity.
\end{itemize}

The idea behind \ourloss\ is that the constraints imposed by the two uni-modal objectives help the standard cross-modal contrastive objective, applying semantic constraints to the common space and, in turn, helping the generalization to different scenarios also in cases of data scarcity.
In the following paragraphs, we will provide better details on the cross- and uni-modal components of the proposed \ourloss\ loss function.

\subsubsection{Cross-modal Contrastive Objective}
The main optimization objective is the one that forces $\mathbf{t}_i$ and $\mathbf{m}_j$ to have maximum cosine similarity when $i = j$.  Specifically, we employ the well-known InfoNCE loss, introduced for cross-modal matching in \cite{zhang2020contrastive} and used, for example, in text-image matching~\cite{RKHRGASAMCKS21-CLIP} or text-motion retrieval~\cite{PBV23-TMR,MSFR23-sigir}. InfoNCE is defined as:
\begin{equation}
\begin{split}
    \mathcal{L}_{\text{nce}} = - \frac{1}{B} \sum_i^B \log \frac{\exp (s(\mathbf{m}_i,\mathbf{t}_i)/\tau)}{\sum_j^B \exp(s(\mathbf{m}_i,\mathbf{t}_j)/\tau)} + \log \frac{\exp (s(\mathbf{m}_i,\mathbf{t}_i)/\tau)}{\sum_j^B \exp(s(\mathbf{m}_j,\mathbf{t}_i)/\tau)},
\end{split}
\end{equation}
where $\tau$ is a temperature parameter learned during training,  $s(\cdot,\cdot)$ is the cosine similarity between two feature vectors, and $B$ is the batch size. 

\subsubsection{Uni-modal Similarity Objective}
Enforcing the cross-modal contrastive objective alone does not directly imply any strong semantic constraints between textual descriptions or motions. This means that if $\mathbf{t}_i$ is close to $\mathbf{m}_j$ in the common space, this does not directly imply that $\mathbf{t}_i$ is close to $\mathbf{t}_j$, and in the same way we cannot assume that $\mathbf{m}_i$ is close to $\mathbf{m}_j$.
This objective serves exactly to bridge this semantic gap in the two uni-modal domains.

We propose the constraint to align the score distribution of cross-modal matching to the two uni-modal domains. We employ the Kullback-Leibler (KL) divergence objective to estimate the distance between these score distributions. Specifically, given the distributions of the scores $\mathcal{S}$ directly derived from the cross-modal and uni-modal cosine similarities:
\begin{equation}
\begin{split}
\mathcal{S}_j^{\text{t2m}}&=\text{softmax}_i\; s(\mathbf{t_j},\mathbf{m_i}) \\
\mathcal{S}_j^{\text{m2t}}&=\text{softmax}_i\; s(\mathbf{t_i}, \mathbf{m_j}) \\
\mathcal{S}_j^{\text{m2m}}&=\text{softmax}_i\; s(\mathbf{m_i}, \mathbf{m_j}) \\
\mathcal{S}_j^{\text{t2t}}&=\text{softmax}_i\; s(\mathbf{t_i}, \mathbf{t_j}) 
\end{split}
\end{equation}
we define the following objectives:
\begin{equation}
\begin{split}
\mathcal{L}_{\text{cross-to-m2m}} &= \frac{1}{B} \sum_j^B\frac{\text{SymmKL}(\mathcal{S}_j^\text{t2m}, \mathcal{S}_j^{\text{m2m}}) + \text{SymmKL}(\mathcal{S}_j^\text{m2t}, \mathcal{S}_j^\text{m2m})}{2} \\
\mathcal{L}_{\text{cross-to-t2t}} &= \frac{1}{B} \sum_j^B\frac{\text{SymmKL}(\mathcal{S}_j^\text{t2m}, \mathcal{S}_j^\text{t2t}) + \text{SymmKL}(\mathcal{S}_j^\text{m2t}, \mathcal{S}_j^\text{t2t})}{2},
\end{split}
\end{equation}
where $\text{SymmKL}(\mathcal{X}, \mathcal{Y}) = \frac{\text{KL}(\mathcal{X}, \mathcal{Y}) + \text{KL}(\mathcal{Y}, \mathcal{X})}{2}$ is the symmetric version of the KL divergence. We use the symmetric version since there is no distribution among the involved ones that can be elected as the \textit{reference} distribution, given that all these distributions can be mutually adjusted during the training phase.

Therefore, we can derive the following objective, which takes into account the score distribution similarity between cross- and uni-modal features:
\begin{equation}
\mathcal{L}_{\text{cross-to-uni}} = \mathcal{L}_{\text{cross-to-t2t}} + \mathcal{L}_{\text{cross-to-m2m}}.
\end{equation}

However, this objective alone may be insufficient. In fact, there is an important training signal that is required to avoid degenerating into potentially trivial or semantically incorrect solutions. In particular, we need, at least in the warmup training phase, some external supervision providing us with meaningful uni-modal score distributions $\mathcal{S}^\text{t2t}$ and $\mathcal{S}^\text{m2m}$ to better guide the organization of motion or text features within the respective uni-modal manifolds in the common space.

To this aim, we employ teacher models able to provide guidance for $\mathcal{S}^\text{t2t}$ and $\mathcal{S}^\text{m2m}$. For the text, we can easily employ a textual model $\mathcal{E}_t^\text{ref}$ trained to estimate the similarity between two sentences. Consequently, the text model works as a \textit{teacher} model that guides the uni-modal scores distribution. Concerning motions, the situation is more challenging, given that motion classifiers~\cite{CCLRL24} or motion autoencoder networks~\cite{SCA23-ecir} able to derive meaningful comparable motion representations are usually trained on different skeleton formats and with diverse label distributions, which makes them not directly transferable to the datasets employed in this work.
A naive yet reasonable solution to this problem is to assume that two motions are similar if their text descriptions are similar. 
In this way, if we assume $\mathcal{S}^\text{textGT} = \text{softmax}_i\; \mathcal{E}_t^\text{ref} (T_i, T_j)$ to be the reliable scores distribution provided by the teacher text model, we can enforce the following:
\begin{align}
    \mathcal{L}^\text{teacher-to-t2t} &= \text{KL}(\mathcal{S}^\text{textGT}, \mathcal{S}^\text{t2t}) \\
    \mathcal{L}^\text{teacher-to-m2m} &= \text{KL}(\mathcal{S}^\text{textGT}, \mathcal{S}^\text{m2m}),
\end{align}
where, at this time, we employ the standard non-symmetric KL loss since $\mathcal{S}^\text{textGT}$ is considered as the true reference distribution.
Therefore, we can derive the following objective, which takes into consideration the distribution gap between an optimal teacher and the uni-modal distributions:

\begin{align}
    \mathcal{L}^\text{teacher-to-uni} = \mathcal{L}^\text{teacher-vs-t2t} + \mathcal{L}^\text{teacher-to-m2m}.
\end{align}

\subsubsection{Final objective}
We finally combine the previously introduced loss functions:
\begin{equation}
    \mathcal{L} = \mathcal{L}_{\text{nce}} + \lambda \mathcal{L}_{\text{cross-to-uni}} + (1 - \lambda) \mathcal{L}^\text{teacher-to-uni},
\end{equation}
where $\lambda$ is a hyper-parameter that balances the contribution of the teacher scores (supervised scores distillation) with respect to the self-alignment between cross- and uni-modal scores (self-sustained scores distillation).
The overall optimization methodology is rooted in the idea that in the beginning, the text teacher's supervision signal helps guide the uni-modal manifolds toward an appropriate configuration, while it becomes unnecessary (or even counterproductive) if it is kept active for the whole training duration. For this reason, we introduce a simple linear scheduling policy for $\lambda$ that swipes this parameter from 0 (full teacher supervision) to 1 (full self-sustained learning regime) across various training epochs, following the swipe function:
\begin{align}
\lambda(t) = \text{clamp}_{[0, 1]}\left( \frac{t - t_\text{start}}{t_\text{end} - t_\text{start}}\right),
\end{align}
where $\text{clamp}_{[0, 1]}(\cdot)$ ensures that the output stays bounded between 0 and 1, $t$ is the current epoch, and $t_\text{start}$ and $t_\text{end}$ are the epochs at which the swipe starts and ends, respectively. We experimentally found that $t_\text{start}=40$ and $t_\text{end}=100$ obtain an optimal performance. Therefore, if not explicitly mentioned, the experiments have been performed with this configuration. However, we dedicated a paragraph in the ablation study (Section~\ref{sec:results-ablation}) to better explore the role of these hyper-parameters.

\section{Experimental Evaluation}

In this section, we briefly describe benchmark datasets and a methodology for the evaluation of text-to-motion retrieval, as used in state-of-the-art approaches. We then report the performance of the proposed joint-dataset and cross-dataset learning and compare the results to prior work. We also present the ablation study of our approach to measure the effects of joint-dataset learning, the usage of different loss functions, and the selection of particular hyper-parameters. In addition, we provide visualization of selected results of text-to-motion retrieval to highlight the strengths/weaknesses of the proposed approach.

\subsection{Datasets and Evaluation Methodology}

We employ HumanML3D~\cite{GZZWJLC22-HumanML3D} and KIT Motion Language~\cite{Plappert2016-KITdataset} datasets that are widely used for 3D text-to-motion generation/retrieval. They provide skeleton data captured by optical marker-based motion capture technologies and several human-written textual descriptions for each motion.
Since the dataset body models are slightly different, we employ the pre-processing pipeline\footnote{\url{https://github.com/Mathux/AMASS-Annotation-Unifier}} which unifies skeleton data into a common motion-text representation. This pipeline also includes extracting rotation-invariant forward kinematics motion features, such as 3D rotations, velocities, or foot contacts, originally proposed in~\cite{GZZWJLC22-HumanML3D}. This unification allows us to evaluate joint-dataset and cross-dataset learning scenarios.

\subsubsection*{KIT Motion-Language Dataset (KITML)} It contains 3,911 recordings of full-body motion in the Master Motor Map form~\cite{terlemez2014master}, along with textual descriptions for each motion. It has a total of 6,278 annotations in English, where each motion recording has one or more annotations that explain the action, like \textit{"A human walks two steps forwards, pivots 180 degrees, and walks two steps back".} 

\subsubsection*{HumanML3D} It is essentially very similar to KITML. However, it is a more recent dataset developed by adding textual annotations to already-existing and widely-used motion-capture datasets -- AMASS~\cite{mahmood2019amass} and HumanAct12~\cite{guo2020action2motion}. It contains 14,616 motions annotated by 44,970 textual descriptions. 

\subsubsection*{Evaluation Methodology}

We employ the same evaluation methodology introduced in~\cite{PBV23-TMR}. In particular, \emph{recall} at rank $k$ ($R@k\uparrow$) measures the percentage of times the correct label is among the top $k$ results, i.e., the higher, the better. Since the recall is evaluated for $k \in \{1,2,3,5,10\}$, the $Rsum$ metric is additionally presented as the sum of recall values over individual settings of $k$. We also report \emph{median rank} ($MedR\downarrow$), i.e., the lower, the better. The results are reported on the test set of the respective datasets, i.e., on unseen motions.
As in~\cite{PBV23-TMR}, we evaluate not only \emph{text-to-motion} retrieval but also the orthogonal task of \emph{motion-to-text} retrieval.
We report the results using the four evaluation protocols also employed in ~\cite{PBV23-TMR}: (i) ``All'', most similar to the protocol employed in \cite{MSFR23-sigir}, where all the text and motions are used as query and retrieval set; (ii) ``All with threshold'', similar to ``All'' but a motion is considered correct if its description matches the query text above a threshold (set to 0.95 as in previous works); (iii) ``Dissimilar subset'', where a subset of 100 text-motion pairs are chosen so that the distances among the sampled texts are maximized; (iv) ``Small batches'', which randomly selects batches of 32 text-motion pairs and reports their average performance. To give an overall assessment of the probed methods, we also report the average of all the metrics over different protocols in the last group of rows in the tables. Notice that, unless otherwise stated, the line ``\ourarchitecture{}`` in the result tables also integrates the proposed \ourloss{} loss function, with $t_\text{start}=40$ and $t_\text{end}=100$ (see \ref{sec:results-ablation} for detailed ablations on these hyper-parameters).


\subsubsection*{Implementation Details}
All the methods were evaluated on three independent runs, and their average values were reported. Concerning TMR, the original code was used to generate the results on the cross-dataset evaluation and on the joint-dataset learning setups. All the methods have been trained for 250 epochs, with a learning rate of 5e-5, employing an RTX 2080Ti GPU. The text encoder, ACTORStyleEncoder, is the same as in TMR and was re-trained from scratch at every run. As the teacher text model $\mathcal{E}_t^\text{ref}$, we employed a pre-trained MPNet \cite{song2020mpnet}. Concerning \ourarchitecture, we downsampled motions to 200 frames if longer than this amount. We employed 1024-D fully connected connections within the attention blocks with 4 heads. As previously mentioned, we used a factorized encoder configuration, with two spatial layers followed by two temporal ones. As in previous works \cite{MSFR23-sigir,PBV23-TMR}, we set the size of the common embedding space to 256.





  

\begin{table}
\tabcolsep 1pt
\caption{\textbf{Joint-dataset} learning: training on KITML+HumanML3D, testing on KITML (the TMR method is also evaluated by training only on KITML: see ``--'' option in the joint-dataset learning (JDL) column).}
\label{tab:resultsJDL-KITeval}
\begin{tabular}{llc|cccccc|cccccc|c}
\toprule
\multirow[c]{2}{*}{\textbf{Protocol}} & \multirow[c]{2}{*}{\textbf{Method}} & \multirow[c]{2}{*}{\textbf{JDL}} & \multicolumn{6}{c|}{Motion-to-text retrieval} & \multicolumn{6}{c|}{Text-to-motion retrieval} & \\
& &  & MedR$\downarrow$ & R@1$\uparrow$ & R@2$\uparrow$ & R@3$\uparrow$ & R@5$\uparrow$ & R@10$\uparrow$ & MedR$\downarrow$ & R@1$\uparrow$ & R@2$\uparrow$ & R@3$\uparrow$ & R@5$\uparrow$ & R@10$\uparrow$ & Rsum$\uparrow$ \\
\toprule
\multirow[c]{4}{*}{\makecell[l]{(a) All}}
 & TMR~\cite{PBV23-TMR} & -- & 15.83	& 10.47 &	13.82 &	21.25 &	29.01 &	40.67 &	16.67 &	6.32 &	11.87 &	17.69 &	26.25 &	39.35 &	216.70 \\
 & TMR~\cite{PBV23-TMR} & \checkmark & 11.33 & 13.78 & 18.07 & 26.68 & 36.43 & 49.15 & 10.33 & 9.37 & 18.24 & 25.15 & 35.37 & 49.79 & 282.03 \\
 & MoT~\cite{MSFR23-sigir} & \checkmark & 11.83 & 12.25 & 15.99 & 25.32 & 34.10 & 48.26 & 10.67 & 8.99 & 17.60 & 24.17 & 33.50 & 48.90 & 269.08 \\
 & MoT++ & \checkmark & 11.50 & 14.42 & 17.94 & 26.63 & 35.54 & 48.60 & 10.83 & 8.99 & 17.81 & 24.39 & 34.73 & 49.83 & 278.88 \\
\cline{1-16}
\multirow[c]{4}{*}{\makecell[l]{(b) All with\\threshold}} & TMR~\cite{PBV23-TMR} & -- & 9.50 &	18.91 &	23.58 &	31.76 &	40.16 &	51.61 &	6.00 &	22.43 &	29.94 &	38.17 &	47.37 &	59.67 &	363.60 \\
 & TMR~\cite{PBV23-TMR} & \checkmark & 6.42 & 23.24 & 28.75 & 38.93 & 48.60 & 62.47 & 4.00 & 27.02 & 37.45 & 48.47 & 58.06 & 70.57 & 443.56 \\
 & MoT~\cite{MSFR23-sigir} & \checkmark & 6.75 & 21.50 & 27.99 & 37.96 & 47.12 & 60.56 & 4.17 & 27.99 & 36.43 & 46.61 & 57.13 & 70.78 & 434.07 \\
 & MoT++ & \checkmark & 6.67 & 21.84 & 27.18 & 38.04 & 47.50 & 62.09 & 4.00 & 24.34 & 37.83 & 47.63 & 57.89 & 72.39 & 436.73 \\
 \cline{1-16}
 \multirow[c]{4}{*}{\makecell[l]{(c) Dissim.\\subset}}
 & TMR~\cite{PBV23-TMR} & -- & 3.42 &	35.03 &	45.24 &	57.48 &	65.65 &	79.93 &	3.33 &	25.17 &	42.18 &	50.68 &	63.61 &	79.25 &	544.22 \\
 & TMR~\cite{PBV23-TMR} & \checkmark & 2.42 & 44.22 & 54.76 & 68.37 & 75.85 & 85.71 & 2.67 & 28.57 & 48.64 & 59.52 & 73.47 & 86.05 & 625.16 \\
 & MoT~\cite{MSFR23-sigir} & \checkmark & 2.58 & 41.50 & 53.74 & 65.65 & 76.19 & 86.05 & 2.67 & 28.91 & 49.66 & 61.56 & 79.25 & 91.16 & 633.67 \\
 & MoT++ & \checkmark & 2.08 & 48.30 & 58.16 & 68.03 & 77.55 & 88.10 & 2.33 & 28.23 & 50.34 & 60.88 & 77.21 & 90.14 & 646.94 \\
\cline{1-16}
\multirow[c]{4}{*}{\makecell[l]{(d) Small\\batches}}
 & TMR~\cite{PBV23-TMR} & -- & 1.41 &	52.91 &	72.05 &	81.68 &	90.71 &	96.09 &	1.45 &	52.04 &	72.22 &	81.60 &	90.63 &	95.87 &	785.80 \\
 & TMR~\cite{PBV23-TMR} & \checkmark & 1.16 & 59.29 & 80.34 & 87.46 & 93.53 & 96.83 & 1.15 & 59.42 & 79.86 & 87.50 & 94.01 & 97.18 & 835.42 \\
 & MoT~\cite{MSFR23-sigir} & \checkmark & 1.23 & 57.12 & 77.12 & 86.03 & 92.88 & 96.92 & 1.10 & 60.76 & 80.51 & 88.71 & 94.88 & 98.05 & 832.98 \\
 & MoT++ & \checkmark & 1.18 & 59.99 & 79.73 & 87.20 & 93.71 & 97.44 & 1.12 & 59.24 & 80.04 & 88.58 & 95.57 & 98.35 & 839.85 \\
\bottomrule
\multirow[c]{4}{*}{\makecell[l]{\bfseries average\\\bfseries over\\\bfseries protocols}}
 & TMR~\cite{PBV23-TMR} & -- & 7.54 &	29.33 &	38.67 &	48.04 &	56.38 &	67.08 &	6.86 &	26.49 &	39.05 &	47.03 &	56.96 &	68.54 &	477.57 \\
 & TMR~\cite{PBV23-TMR} & \checkmark & \textbf{5.33} & 35.13 & 45.48 & \textbf{55.36} & \textbf{63.60} & 73.54 & \textbf{4.54} & 31.10 & 46.05 & 55.16 & 65.23 & 75.90 & 546.55 \\
 & MoT~\cite{MSFR23-sigir} & \checkmark & 5.60 & 33.09 & 43.71 & 53.74 & 62.57 & 72.95 & 4.65 & \textbf{31.66} & 46.05 & 55.26 & 66.19 & 77.22 & 542.44 \\
 & MoT++ & \checkmark & 5.36 & \textbf{36.14} & \textbf{45.75} & 54.98 & 63.57 & \textbf{74.06} & 4.57 & 30.20 & \textbf{46.50} & \textbf{55.37} & \textbf{66.35} & \textbf{77.68} & \bfseries 550.60 \\
\bottomrule
\end{tabular}
\end{table}

\begin{table}
\tabcolsep 1pt
\caption{\textbf{Joint-dataset} learning: training on KITML+HumanML3D, testing on HumanML3D (the TMR method is also evaluated by training only on HumanML3D: see ``--'' option in the joint-dataset learning (JDL) column).}
\label{tab:resultsJDL-HumanML3Deval}
\begin{tabular}{llc|cccccc|cccccc|c}
\toprule
\multirow[c]{2}{*}{\textbf{Protocol}} & \multirow[c]{2}{*}{\textbf{Method}} & \multirow[c]{2}{*}{\textbf{JDL}} & \multicolumn{6}{c|}{Motion-to-text retrieval} & \multicolumn{6}{c|}{Text-to-motion retrieval} & \\
& &  & MedR$\downarrow$ & R@1$\uparrow$ & R@2$\uparrow$ & R@3$\uparrow$ & R@5$\uparrow$ & R@10$\uparrow$ & MedR$\downarrow$ & R@1$\uparrow$ & R@2$\uparrow$ & R@3$\uparrow$ & R@5$\uparrow$ & R@10$\uparrow$ & Rsum$\uparrow$ \\
\toprule
\multirow[c]{4}{*}{\makecell[l]{(a) All}}
 & TMR~\cite{PBV23-TMR} & -- & 28.17 & 8.72 & 11.07 & 15.85 & 21.35 & 31.20 & 27.67 & 5.35 & 10.11 & 13.45 & 19.33 & 30.74 & 167.17 \\
 & TMR~\cite{PBV23-TMR} & \checkmark & 27.33 & 8.75 & 10.96 & 16.40 & 21.93 & 31.61 & 27.67 & 5.48 & 10.17 & 13.53 & 19.26 & 30.40 & 168.49 \\
 & MoT~\cite{MSFR23-sigir} & \checkmark & 35.83 & 6.41 & 8.26 & 12.61 & 17.24 & 26.23 & 34.67 & 4.27 & 7.92 & 10.83 & 16.03 & 25.91 & 135.71 \\
 & MoT++ & \checkmark & 25.92 & 8.79 & 11.45 & 17.00 & 22.67 & 32.22 & 26.33 & 5.24 & 10.00 & 14.07 & 20.28 & 31.01 & 172.73 \\
\cline{1-16}
\multirow[c]{4}{*}{\makecell[l]{(b) All with\\threshold}}
 & TMR~\cite{PBV23-TMR} & -- & 22.00 & 12.23 & 14.39 & 20.19 & 26.11 & 36.26 & 18.33 & 11.42 & 15.67 & 21.19 & 27.71 & 39.47 & 224.64 \\
 & TMR~\cite{PBV23-TMR} & \checkmark & 21.00 & 12.24 & 14.43 & 20.75 & 26.77 & 36.93 & 17.33 & 11.38 & 15.73 & 21.04 & 28.03 & 39.85 & 227.15 \\
 & MoT~\cite{MSFR23-sigir} & \checkmark & 28.33 & 10.08 & 12.20 & 17.49 & 22.73 & 31.87 & 21.67 & 9.36 & 13.53 & 18.29 & 24.88 & 36.25 & 196.68 \\
 & MoT++ & \checkmark & 20.75 & 11.57 & 14.09 & 20.47 & 26.67 & 36.73 & 17.00 & 10.83 & 15.38 & 21.28 & 29.03 & 40.58 & 226.63 \\
\cline{1-16}
\multirow[c]{4}{*}{\makecell[l]{(c) Dissim.\\subset}}
 & TMR~\cite{PBV23-TMR} & -- & 1.83 & 49.33 & 67.00 & 73.33 & 81.00 & 88.67 & 2.00 & 47.00 & 67.00 & 72.33 & 80.00 & 88.00 & 713.66 \\
 & TMR~\cite{PBV23-TMR} & \checkmark & 2.00 & 46.33 & 67.67 & 73.67 & 83.00 & 88.33 & 2.00 & 45.67 & 65.67 & 74.00 & 82.00 & 88.67 & 715.01 \\
 & MoT~\cite{MSFR23-sigir} & \checkmark & 2.00 & 42.67 & 61.00 & 71.67 & 80.00 & 89.33 & 2.00 & 42.33 & 61.33 & 72.33 & 82.33 & 90.00 & 692.99 \\
 & MoT++ & \checkmark & 1.83 & 48.33 & 64.33 & 74.00 & 82.33 & 90.33 & 1.83 & 49.00 & 66.67 & 72.33 & 80.33 & 89.00 & 716.65 \\
\cline{1-16}
\multirow[c]{4}{*}{\makecell[l]{(d) Small\\batches}}
 & TMR~\cite{PBV23-TMR} & -- & 1.02 & 68.03 & 82.38 & 87.86 & 92.32 & 96.27 & 1.01 & 67.32 & 82.00 & 87.43 & 92.22 & 96.36 & 852.19 \\
 & TMR~\cite{PBV23-TMR} & \checkmark & 1.02 & 68.45 & 82.89 & 88.26 & 92.74 & 96.47 & 1.02 & 68.45 & 82.71 & 87.94 & 92.52 & 96.40 & 856.83 \\
 & MoT~\cite{MSFR23-sigir} & \checkmark & 1.04 & 64.83 & 80.47 & 86.69 & 92.35 & 96.83 & 1.04 & 64.99 & 80.51 & 86.91 & 91.91 & 96.59 & 842.08 \\
 & MoT++ & \checkmark & 1.01 & 69.02 & 83.71 & 89.27 & 93.71 & 97.35 & 1.01 & 68.43 & 82.74 & 88.69 & 93.41 & 97.31 & 863.64 \\
\bottomrule
\multirow[c]{4}{*}{\makecell[l]{\bfseries average\\\bfseries over\\\bfseries protocols}}
 & TMR~\cite{PBV23-TMR} & -- & 13.26 & \textbf{34.58} & 43.71 & 49.31 & 55.19 & 63.10 & 12.25 & 32.77 & 43.69 & 48.60 & 54.82 & 63.64 & 489.41 \\
 & TMR~\cite{PBV23-TMR} & \checkmark & 12.84 & 33.94 & \textbf{43.99} & 49.77 & 56.11 & 63.34 & 12.00 & 32.74 & 43.57 & \textbf{49.13} & 55.45 & 63.83 & 491.87 \\
 & MoT~\cite{MSFR23-sigir} & \checkmark & 16.80 & 31.00 & 40.48 & 47.12 & 53.08 & 61.06 & 14.84 & 30.24 & 40.83 & 47.09 & 53.79 & 62.19 & 466.88 \\
 & MoT++ & \checkmark & \textbf{12.38} & 34.43 & 43.40 & \textbf{50.18} & \textbf{56.35} & \textbf{64.16} & \textbf{11.54} & \textbf{33.38} & \textbf{43.70} & 49.09 & \textbf{55.76} & \textbf{64.47} & \bfseries 494.92 \\
\bottomrule
\end{tabular}
\end{table}

\subsection{Results of Joint-Dataset and Cross-Dataset Learning}

We first analyze the effect of \emph{joint-dataset} learning (JDL).
Table~\ref{tab:resultsJDL-KITeval} reports the results of selected methods evaluated on KITML as the test dataset. The JDL (\textit{joint dataset learning}) column indicates whether both KITML and HumanML3D datasets were used simultaneously for training (``\checkmark'' option), or just the KITML dataset was utilized (``--'' option). When focusing on the last table section (i.e., ``average over protocols''), we can observe the clearly best result (Rsum 550.60) of the proposed \ourarchitecture{} method compared to the result (Rsum 477.57) of the state-of-the-art TMR method~\cite{PBV23-TMR}. To be fair, we also evaluate the provided TMR\footnote{\url{https://github.com/Mathux/TMR}} and MoT\footnote{\url{https://github.com/mesnico/text-to-motion-retrieval}} implementations using the joint-dataset learning approach -- we can still observe that our \ourarchitecture{} outperforms both these approaches on average(Rsum 550.60 vs 546.55), noticeably achieving an average improvement of around 0.7\% in R@10 on motion-to-text and 1.7\% on text-to-motion retrieval scenarios with respect TMR, only obtaining slightly worse results on "All" and "All with threshold" protocols. A similar trend can be observed in Table~\ref{tab:resultsJDL-HumanML3Deval} where HumanML3D is used as the test dataset. Again, the proposed \ourarchitecture{} approach outperforms existing approaches, with an increase on the R@10 metric of 1.2\% on motion-to-text and of 1.0\% on text-to-motion retrieval, also obtaining the best median rank (MedR) values. Interestingly, joint-dataset learning seems not to help TMR trained purely on HumanML3D too much since KITML is quite a small and motion-specific dataset that provides limited generalizability for HumanML3D.

To also have a better understanding of the generalization abilities of the proposed models and the explored baselines, we also evaluate the \emph{cross-dataset} scenario by training on HumanML3D and testing on KITML (the opposite variant, i.e., training on KITML and testing on HumanML3D, is less meaningful due to the much richer nature of HumanML3D). The results in Table~\ref{tab:resultsHumanML3Dtrain-KITeval} demonstrate that \ourarchitecture{} again reaches the best result on average (Rsum 527.4) in comparison with the results of TMR or MoT (Rsum $\sim$516.5), with an increase on R@10 metric of 1.5\% on motion-to-text and of 2.0\% on text-to-motion with respect MoT, which noticeably obtains better results with respect to TMR. Therefore, these results also suggest that the two MoT-based motion encoders are more prone to generalize across diverse datasets. Another interesting observation is that the result of \ourarchitecture{} (Rsum 527.4) is much better than the result of TMR trained purposely on KITML (Rsum 477.6 in Table~\ref{tab:resultsJDL-KITeval}), which proves the ability of \ourarchitecture{} to generalize well on unseen data.

\begin{figure}[t]
 \centering
 \begin{subfigure}[t]{.42\linewidth}
     \centering
     \includegraphics[width=\linewidth]{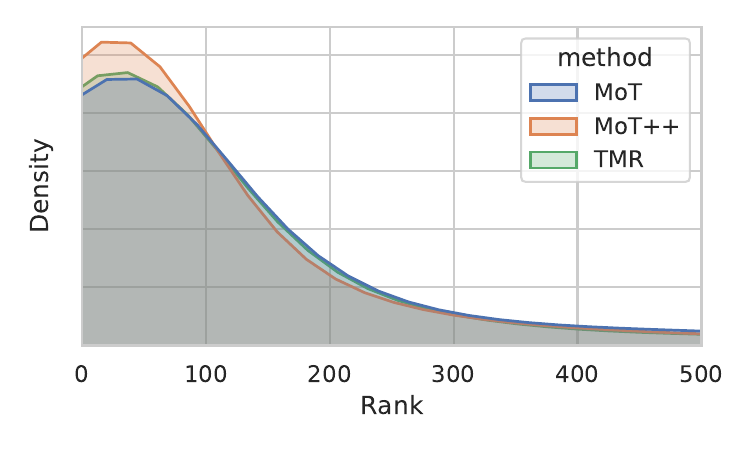}
     \caption{Text-to-motion rank distribution}
     \label{fig:t2m-r10-vs-dim}
 \end{subfigure}
 \begin{subfigure}[t]{.42\linewidth}
     \centering
     \includegraphics[width=\linewidth]{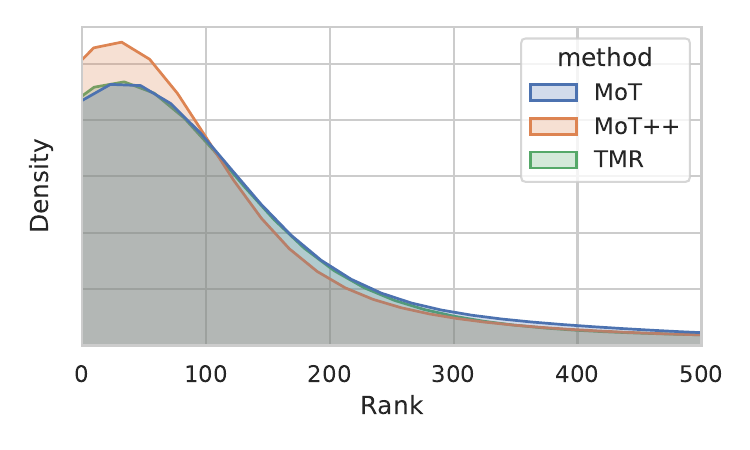}
     \caption{Motion-to-text rank distribution}
     \label{fig:m2t-r10-vs-dim}
 \end{subfigure}
 \caption{Distributions of ranks ($x$-axis) of relevant objects retrieved for (a) text-to-motion and (b) motion-to-text scenarios using joint-dataset learning: training on KITML+HumanML3D, testing on HumanML3D.}
 \label{fig:results-ranks}
 \end{figure}

\begin{table}
\tabcolsep 1pt
\caption{\textbf{Cross-dataset} inference: training on HumanML3D, testing on KITML.}
\label{tab:resultsHumanML3Dtrain-KITeval}
\begin{tabular}{ll|cccccc|cccccc|c}
\toprule
\multirow[c]{2}{*}{\textbf{Protocol}} & \multirow[c]{2}{*}{\textbf{Method}} & \multicolumn{6}{c|}{Motion-to-text retrieval} & \multicolumn{6}{c|}{Text-to-motion retrieval} & \\
& & MedR$\downarrow$ & R@1$\uparrow$ & R@2$\uparrow$ & R@3$\uparrow$ & R@5$\uparrow$ & R@10$\uparrow$ & MedR$\downarrow$ & R@1$\uparrow$ & R@2$\uparrow$ & R@3$\uparrow$ & R@5$\uparrow$ & R@10$\uparrow$ & Rsum$\uparrow$ \\
\toprule
\multirow[c]{3}{*}{\makecell[l]{(a) All}}
 & TMR~\cite{PBV23-TMR} & 12.33 & 12.80 & 15.31 & 24.21 & 33.25 & 46.48 & 13.33 & 8.10 & 15.65 & 21.88 & 31.00 & 44.87 & 253.55 \\
 & MoT~\cite{MSFR23-sigir} & 13.42 & 12.38 & 15.40 & 22.65 & 30.57 & 45.21 & 15.17 & 7.12 & 14.21 & 19.59 & 28.37 & 41.52 & 237.02 \\
 & MoT++ & 12.00 & 12.47 & 15.95 & 24.13 & 33.50 & 46.95 & 12.83 & 7.17 & 14.38 & 20.44 & 29.94 & 46.02 & 250.95 \\
\cline{1-15}
\multirow[c]{3}{*}{\makecell[l]{(b) All with\\threshold}}
 & TMR~\cite{PBV23-TMR} & 7.33 & 22.94 & 26.63 & 38.34 & 46.10 & 59.12 & 5.00 & 25.83 & 33.59 & 42.45 & 51.65 & 64.46 & 411.11 \\
 & MoT~\cite{MSFR23-sigir} & 7.83 & 21.50 & 26.25 & 35.03 & 43.00 & 56.53 & 5.33 & 21.97 & 32.06 & 38.76 & 49.70 & 63.99 & 388.79 \\
 & MoT++ & 7.00 & 20.99 & 26.76 & 36.77 & 47.12 & 59.84 & 5.00 & 22.86 & 32.19 & 43.34 & 53.01 & 67.35 & 410.23 \\
\cline{1-15}
\multirow[c]{3}{*}{\makecell[l]{(c) Dissim.\\subset}}
 & TMR~\cite{PBV23-TMR} & 3.00 & 36.05 & 47.96 & 64.63 & 73.13 & 83.33 & 3.00 & 26.19 & 44.56 & 57.14 & 69.05 & 82.65 & 584.69 \\
 & MoT~\cite{MSFR23-sigir} & 2.67 & 39.80 & 50.68 & 64.63 & 78.57 & 88.09 & 2.17 & 32.65 & 52.38 & 65.31 & 78.23 & 89.80 & 640.14 \\
 & MoT++ & 2.58 & 41.84 & 53.06 & 66.67 & 77.21 & 88.10 & 2.67 & 27.89 & 49.66 & 61.90 & 74.15 & 88.78 & 629.26 \\
\cline{1-15}
\multirow[c]{3}{*}{\makecell[l]{(d) Small\\batches}}
 & TMR~\cite{PBV23-TMR} & 1.25 & 56.77 & 75.74 & 85.24 & 93.36 & 97.88 & 1.26 & 55.43 & 75.52 & 85.98 & 93.32 & 97.70 & 816.94 \\
 & MoT~\cite{MSFR23-sigir} & 1.31 & 54.30 & 73.87 & 82.64 & 92.67 & 98.01 & 1.40 & 52.26 & 72.74 & 82.59 & 92.80 & 98.27 & 800.15 \\
 & MoT++ & 1.19 & 58.51 & 77.69 & 86.15 & 93.19 & 97.53 & 1.24 & 56.03 & 75.09 & 84.29 & 93.06 & 97.70 & 819.24 \\
\bottomrule
\multirow[c]{3}{*}{\makecell[l]{\bfseries average\\\bfseries over\\\bfseries protocols}}
 & TMR~\cite{PBV23-TMR} & 5.98 & 32.14 & 41.41 & 53.10 & 61.46 & 71.70 & 5.65 & \textbf{28.89} & 42.33 & 51.86 & 61.26 & 72.42 & 516.57 \\
 & MoT~\cite{MSFR23-sigir} & 6.31 & 32.00 & 41.55 & 51.24 & 61.20 & 71.96 & 6.02 & 28.50 & \textbf{42.85} & 51.56 & 62.28 & 73.40 & 516.54 \\
 & MoT++ & \textbf{5.69} & \textbf{33.45} & \textbf{43.36} & \textbf{53.43} & \textbf{62.75} & \textbf{73.10} & \textbf{5.44} & 28.49 & 42.83 & \textbf{52.49} & \textbf{62.54} & \textbf{74.96} & \bfseries 527.40 \\
\bottomrule
\end{tabular}
\end{table}

To further show the positive effect of \ourloss\ and joint dataset training on the retrieval results, we plot the distribution of the ranks -- positions of the first relevant objects retrieved -- in Fig.~\ref{fig:results-ranks} for all the test queries, on HumanML3D. We can observe that \ourarchitecture{} has clearly more queries answered with a better quality -- a query match is located within the $\sim$50 nearest neighbors in roughly about 15--20\,\% more queries.
In summary, joint-dataset learning enhances the performance of all methods across various retrieval tasks and protocols. \ourarchitecture{} with JDL generally provides the best performance, especially in terms of ``Rsum'', making it a robust choice for motion-to-text and text-to-motion retrieval tasks. 

\subsection{Ablation Study}
\label{sec:results-ablation}

In this section, we are interested in understanding the importance of all the introduced components. Specifically, in (i) understanding whether the joint-dataset learning, together with the proposed \ourloss{} loss, effectively contributes to improving the overall performance in both text-to-motion and motion-to-text scenarios; (ii) understanding the role of the $\lambda$ swipe hyper-parameters introduced in \ourloss; (iii) providing a quantitative evaluation of motion-to-motion retrieval, which is a nice byproduct of \ourloss; and (iv) providing some qualitative results by comparing our outcomes with the ones from the state-of-the-art TMR method.

\subsubsection{Role of \ourloss{} + joint-dataset learning}
We aim to explore the important roles that both the joint-dataset learning and our proposed \ourloss{} objective have on the final model. In Table~\ref{tab:resultsJDL-HumanML3Deval-lossAblations2}, we report the results on \ourarchitecture\ by alternatively enabling or disabling the joint-dataset learning (JDL) and using either InfoNCE with filtering (InfoNCE+F) or \ourloss. As we can notice, the best overall results are obtained with the combination of joint-dataset training and \ourloss, obtaining an overall increase in Rsum between 485.89 and 494.92, with an increase of 3.0\% on R@10 on motion-to-text and 2.2\% on text-to-motion. These results demonstrate the nice effect that the availability of more data and the careful constraints imposed by \ourloss\ on the common space have on the overall model.

\begin{table}
\tabcolsep 1pt
\caption{\ourloss\ ablations on MoT++ using single-dataset learning (training on HumanML3D: ``--'' option in JDL) and joint-dataset learning (training on KITML+HumanML3D: ``\checkmark'' option in JDL), testing on HumanML3D.}
\label{tab:resultsJDL-HumanML3Deval-lossAblations2}
\begin{tabular}{llc|cccccc|cccccc|c}
\toprule
\multirow[c]{2}{*}{\textbf{Protocol}} & \multirow[c]{2}{*}{\makecell[l]{\bfseries Loss\\ \bfseries function}} & \multirow[c]{2}{*}{\textbf{JDL}} & \multicolumn{6}{c|}{Motion-to-text retrieval} & \multicolumn{6}{c|}{Text-to-motion retrieval} & \\
& &  & MedR$\downarrow$ & R@1$\uparrow$ & R@2$\uparrow$ & R@3$\uparrow$ & R@5$\uparrow$ & R@10$\uparrow$ & MedR$\downarrow$ & R@1$\uparrow$ & R@2$\uparrow$ & R@3$\uparrow$ & R@5$\uparrow$ & R@10$\uparrow$ & Rsum$\uparrow$ \\
\toprule
\multirow[c]{4}{*}{\makecell[l]{(a) All}}
 & InfoNCE+F~\cite{PBV23-TMR} & -- & 29.00 & 8.32 & 10.40 & 15.79 & 21.22 & 30.03 & 30.00 & 4.94 & 9.50 & 12.92 & 18.62 & 29.46 & 161.20 \\
 & InfoNCE+F~\cite{PBV23-TMR} & \checkmark & 30.00 & 8.58 & 11.08 & 15.62 & 20.86 & 30.47 & 30.33 & 4.77 & 9.47 & 12.64 & 18.42 & 28.74 & 160.65 \\
 & \ourloss\ & -- & 26.50 & 9.06 & 11.54 & 16.49 & 22.08 & 31.70 & 27.17 & 5.55 & 10.67 & 14.10 & 20.31 & 31.21 & 172.71 \\
 & \ourloss\ & \checkmark & 25.92 & 8.79 & 11.45 & 17.00 & 22.67 & 32.22 & 26.33 & 5.24 & 10.00 & 14.07 & 20.28 & 31.01 & 172.73 \\
\cline{1-16}
\multirow[c]{4}{*}{\makecell[l]{(b) All with\\threshold}}
 & InfoNCE+F~\cite{PBV23-TMR} & -- & 23.00 & 12.00 & 13.77 & 20.04 & 25.92 & 35.08 & 19.33 & 11.14 & 15.00 & 20.50 & 26.96 & 38.18 & 218.59 \\
 & InfoNCE+F~\cite{PBV23-TMR} & \checkmark & 23.00 & 11.92 & 14.26 & 19.71 & 25.52 & 35.64 & 20.00 & 10.52 & 14.53 & 19.96 & 26.47 & 37.77 & 216.30 \\
 & \ourloss\ & -- & 21.33 & 11.89 & 14.00 & 20.02 & 26.19 & 36.24 & 17.67 & 11.24 & 16.05 & 21.45 & 28.52 & 40.05 & 225.65 \\
 & \ourloss\ & \checkmark & 20.75 & 11.57 & 14.09 & 20.47 & 26.67 & 36.73 & 17.00 & 10.83 & 15.38 & 21.28 & 29.03 & 40.58 & 226.63 \\
\cline{1-16}
\multirow[c]{4}{*}{\makecell[l]{(c) Dissim.\\subset}}
 & InfoNCE+F~\cite{PBV23-TMR} & -- & 1.33 & 50.00 & 66.33 & 73.33 & 81.33 & 88.00 & 1.83 & 48.00 & 67.33 & 74.33 & 80.33 & 88.33 & 717.31 \\
 & InfoNCE+F~\cite{PBV23-TMR} & \checkmark & 2.00 & 43.00 & 66.00 & 74.00 & 83.00 & 88.00 & 1.83 & 46.00 & 65.33 & 73.33 & 79.33 & 88.33 & 706.32 \\
 & \ourloss\ & -- & 2.00 & 47.00 & 65.00 & 73.67 & 80.33 & 88.33 & 1.83 & 47.33 & 65.00 & 73.67 & 79.67 & 87.33 & 707.33 \\
 & \ourloss\ & \checkmark & 1.83 & 48.33 & 64.33 & 74.00 & 82.33 & 90.33 & 1.83 & 49.00 & 66.67 & 72.33 & 80.33 & 89.00 & 716.65 \\
\cline{1-16}
\multirow[c]{4}{*}{\makecell[l]{(d) Small\\batches}}
 & InfoNCE+F~\cite{PBV23-TMR} & -- & 1.04 & 67.37 & 81.55 & 87.29 & 91.75 & 95.87 & 1.03 & 66.68 & 81.21 & 86.80 & 91.70 & 96.16 & 846.38 \\
 & InfoNCE+F~\cite{PBV23-TMR} & \checkmark & 1.02 & 67.72 & 82.28 & 87.74 & 92.04 & 95.96 & 1.02 & 66.68 & 81.65 & 87.21 & 91.79 & 95.97 & 849.04 \\
 & \ourloss\ & -- & 1.01 & 68.74 & 82.82 & 88.22 & 93.01 & 97.14 & 1.01 & 68.13 & 82.38 & 87.60 & 92.37 & 97.06 & 857.47 \\
 & \ourloss\ & \checkmark & 1.01 & 69.02 & 83.71 & 89.27 & 93.71 & 97.35 & 1.01 & 68.43 & 82.74 & 88.69 & 93.41 & 97.31 & 863.64 \\
\bottomrule
\multirow[c]{4}{*}{\makecell[l]{\bfseries average\\\bfseries over\\\bfseries protocols}}
 & InfoNCE+F~\cite{PBV23-TMR} & -- & 13.59 & 34.42 & 43.01 & 49.11 & 55.06 & 62.25 & 13.05 & 32.69 & 43.26 & 48.64 & 54.41 & 63.04 & 485.89 \\
 & InfoNCE+F~\cite{PBV23-TMR} & \checkmark & 14.00 & 32.80 & 43.40 & 49.26 & 55.35 & 62.52 & 13.30 & 31.99 & 42.74 & 48.28 & 54.00 & 62.70 & 483.04 \\
 & \ourloss\ & -- & 12.71 & 34.17 & 43.34 & 49.60 & 55.40 & 63.35 & 11.92 & 33.06 & 43.53 & \textbf{49.20} & 55.22 & 63.91 & 490.78 \\
 & \ourloss\ & \checkmark & \textbf{12.38} & \textbf{34.43} & \textbf{43.40} & \textbf{50.18} & \textbf{56.35} & \textbf{64.16} & \textbf{11.54} & \textbf{33.38} & \textbf{43.70} & 49.09 & \textbf{55.76} & \textbf{64.47} & \bfseries 494.92 \\
\bottomrule
\end{tabular}
\end{table}

 \begin{table}
\tabcolsep 1pt
\caption{\ourloss\ ablations on MoT++ using \textbf{joint-dataset} learning: training on KITML+HumanML3D, testing on HumanML3D. We indicate as ``\ourloss\ x-y'' the experiment performed using $t_\text{start}=x$ and $t_\text{end}=y$.}
\label{tab:resultsJDL-HumanML3Deval-lossAblations}
\begin{tabular}{ll|cccccc|cccccc|c}
\toprule
\multirow[c]{2}{*}{\textbf{Protocol}} & \multirow[c]{2}{*}{\bfseries Loss function} & \multicolumn{6}{c|}{Motion-to-text retrieval} & \multicolumn{6}{c|}{Text-to-motion retrieval} & \\
& & MedR$\downarrow$ & R@1$\uparrow$ & R@2$\uparrow$ & R@3$\uparrow$ & R@5$\uparrow$ & R@10$\uparrow$ & MedR$\downarrow$ & R@1$\uparrow$ & R@2$\uparrow$ & R@3$\uparrow$ & R@5$\uparrow$ & R@10$\uparrow$ & Rsum$\uparrow$ \\
\toprule
\multirow[c]{6}{*}{\makecell[l]{(a) All}}
 & InfoNCE+F~\cite{PBV23-TMR} & 30.00 & 8.58 & 11.08 & 15.62 & 20.86 & 30.47 & 30.33 & 4.77 & 9.47 & 12.64 & 18.42 & 28.74 & 160.65 \\
 & \ourloss\ self & 27.67 & 8.46 & 10.93 & 16.13 & 21.79 & 31.84 & 28.00 & 5.26 & 9.84 & 13.25 & 19.44 & 30.06 & 167.00 \\
 & \ourloss\ 40-100 & 25.92 & 8.79 & 11.45 & 17.00 & 22.67 & 32.22 & 26.33 & 5.24 & 10.00 & 14.07 & 20.28 & 31.01 & 172.73 \\
 & \ourloss\ 80-140 & 27.00 & 8.86 & 11.19 & 16.58 & 22.57 & 32.29 & 27.33 & 5.12 & 9.93 & 13.50 & 19.67 & 31.21 & 170.92 \\
 & \ourloss\ 140-200 & 28.17 & 9.15 & 11.60 & 16.61 & 21.84 & 31.27 & 28.33 & 5.33 & 10.35 & 13.96 & 19.94 & 29.99 & 170.04 \\
 & \ourloss\ supervised & 30.50 & 8.45 & 10.85 & 15.77 & 20.99 & 29.88 & 30.67 & 4.76 & 9.39 & 12.80 & 18.77 & 29.53 & 161.19 \\
\cline{1-15}
\multirow[c]{6}{*}{\makecell[l]{(b) All with\\threshold}}
 & InfoNCE+F~\cite{PBV23-TMR} & 23.00 & 11.92 & 14.26 & 19.71 & 25.52 & 35.64 & 20.00 & 10.52 & 14.53 & 19.96 & 26.47 & 37.77 & 216.30 \\
 & \ourloss\ self & 21.58 & 11.34 & 13.53 & 20.00 & 26.00 & 36.61 & 18.00 & 11.06 & 15.25 & 20.56 & 27.76 & 39.33 & 221.44 \\
 & \ourloss\ 40-100 & 20.75 & 11.57 & 14.09 & 20.47 & 26.67 & 36.73 & 17.00 & 10.83 & 15.38 & 21.28 & 29.03 & 40.58 & 226.63 \\
 & \ourloss\ 80-140 & 21.50 & 11.65 & 13.92 & 20.50 & 26.89 & 36.79 & 17.00 & 10.52 & 15.27 & 20.81 & 28.01 & 40.59 & 224.95 \\
 & \ourloss\ 140-200 & 21.67 & 12.11 & 14.39 & 20.20 & 26.09 & 36.07 & 17.67 & 11.31 & 15.74 & 21.07 & 28.11 & 38.92 & 224.01 \\
 & \ourloss\ supervised & 23.33 & 11.52 & 13.85 & 19.53 & 25.47 & 34.89 & 19.33 & 10.19 & 14.85 & 20.01 & 27.15 & 38.38 & 215.84 \\
 \cline{1-15}
\multirow[c]{6}{*}{\makecell[l]{(c) Dissim.\\subset}}
 & InfoNCE+F~\cite{PBV23-TMR} & 2.00 & 43.00 & 66.00 & 74.00 & 83.00 & 88.00 & 1.83 & 46.00 & 65.33 & 73.33 & 79.33 & 88.33 & 706.32 \\
 & \ourloss\ self & 1.67 & 48.00 & 66.33 & 74.67 & 83.33 & 90.67 & 1.67 & 47.33 & 65.33 & 73.67 & 82.67 & 90.33 & 722.33 \\
 & \ourloss\ 40-100 & 1.83 & 48.33 & 64.33 & 74.00 & 82.33 & 90.33 & 1.83 & 49.00 & 66.67 & 72.33 & 80.33 & 89.00 & 716.65 \\
 & \ourloss\ 80-140 & 2.00 & 45.67 & 62.33 & 70.33 & 80.00 & 88.67 & 1.83 & 47.33 & 63.33 & 69.67 & 76.33 & 88.33 & 691.99 \\
 & \ourloss\ 140-200 & 1.83 & 48.00 & 62.33 & 72.33 & 81.33 & 89.00 & 1.67 & 48.67 & 63.67 & 69.67 & 79.67 & 86.67 & 701.34 \\
 & \ourloss\ supervised & 1.67 & 47.67 & 63.33 & 72.67 & 79.67 & 87.33 & 2.00 & 45.67 & 65.00 & 71.33 & 79.00 & 86.67 & 698.34 \\
\cline{1-15}
 \multirow[c]{6}{*}{\makecell[l]{(d) Small\\batches}}
 & InfoNCE+F~\cite{PBV23-TMR} & 1.02 & 67.72 & 82.28 & 87.74 & 92.04 & 95.96 & 1.02 & 66.68 & 81.65 & 87.21 & 91.79 & 95.97 & 849.04 \\
 & \ourloss\ self & 1.00 & 68.51 & 83.12 & 88.69 & 93.57 & 97.39 & 1.02 & 67.66 & 82.24 & 88.17 & 93.34 & 97.31 & 860.00 \\
 & \ourloss\ 40-100 & 1.01 & 69.02 & 83.71 & 89.27 & 93.71 & 97.35 & 1.01 & 68.43 & 82.74 & 88.69 & 93.41 & 97.31 & 863.64 \\
 & \ourloss\ 80-140 & 1.01 & 68.21 & 82.88 & 88.69 & 93.41 & 97.36 & 1.02 & 67.81 & 81.93 & 87.92 & 93.01 & 97.21 & 858.43 \\
 & \ourloss\ 140-200 & 1.01 & 67.57 & 81.99 & 87.70 & 93.24 & 97.27 & 1.01 & 67.01 & 81.45 & 87.17 & 92.70 & 97.16 & 853.26 \\
 & \ourloss\ supervised & 1.02 & 66.04 & 80.67 & 86.52 & 91.99 & 96.67 & 1.03 & 65.78 & 79.65 & 85.71 & 91.52 & 96.38 & 840.93 \\
\bottomrule
\multirow[c]{6}{*}{\makecell[l]{\bfseries average\\\bfseries over\\\bfseries protocols}}
 & InfoNCE+F~\cite{PBV23-TMR} & 14.00 & 32.80 & 43.40 & 49.26 & 55.35 & 62.52 & 13.30 & 31.99 & 42.74 & 48.28 & 54.00 & 62.70 & 483.04 \\
 & \ourloss\ self & 12.98 & 34.08 & \textbf{43.48} & 49.87 & 56.17 & 64.13 & 12.17 & 32.83 & 43.17 & 48.91 & \textbf{55.80} & 64.26 & 492.70 \\
 & \ourloss\ 40-100 & \textbf{12.38} & \textbf{34.43} & 43.40 & \textbf{50.18} & \textbf{56.35} & \textbf{64.16} & \textbf{11.54} & \textbf{33.38} & \textbf{43.70} & \textbf{49.09} & 55.76 & \textbf{64.47} & \bfseries 494.92 \\
 & \ourloss\ 80-140 & 12.88 & 33.60 & 42.58 & 49.03 & 55.72 & 63.78 & 11.80 & 32.70 & 42.61 & 47.97 & 54.26 & 64.34 & 486.59 \\
 & \ourloss\ 140-200 & 13.17 & 34.21 & 42.58 & 49.21 & 55.63 & 63.40 & 12.17 & 33.08 & 42.80 & 47.96 & 55.10 & 63.18 & 487.15 \\
 & \ourloss\ supervised & 14.13 & 33.42 & 42.18 & 48.62 & 54.53 & 62.20 & 13.26 & 31.60 & 42.22 & 47.46 & 54.11 & 62.74 & 479.08 \\
\bottomrule
\end{tabular}
\end{table}

\subsubsection{\ourloss\ hyper-parameters}
Our loss formulation is driven by two main hyper-parameters, $t_\text{start}$ and $t_\text{end}$, which define the start and end epochs of the linear transition between the text supervision enforced by the teacher model $\mathcal{E}_t^\text{ref}$ and the self-sustained regime where the teacher is fully disabled. 
In Table~\ref{tab:resultsJDL-HumanML3Deval-lossAblations}, we report experiments performed on \ourarchitecture, on the joint-dataset learning setup, for different ranges of $t_\text{start}$ and $t_\text{end}$, where we indicate as "\ourloss\ x-y" the experiment performed using $t_\text{start}=x$ and $t_\text{end}=y$. We also report the edge cases in which the text teacher is always disabled (\ourloss\ self) and in which it is instead always active (\ourloss\ supervised). As we can notice, the best result is obtained when $\lambda$ is varied between epochs 40 and 100 (which is the configuration employed in Tables~\ref{tab:resultsJDL-KITeval}, \ref{tab:resultsJDL-HumanML3Deval}, \ref{tab:resultsHumanML3Dtrain-KITeval}). By looking at the final Rsum value over the averaged measures, it is interesting to notice that almost all the configurations (comprising the fully self-sustained one), can surpass the InfoNCE with filtering (InfoNCE+F) employed by TMR in~\cite{PBV23-TMR}. Notably, the fully supervised scenario is the one achieving the worst results. This suggests that, while the supervision signal is helpful in the early training epochs, it is detrimental if kept active until the last iterations. This may be due to the fact that the text-motion domain is much more specific than the general-purpose knowledge conserved within the teacher $\mathcal{E}_t^\text{ref}$. Therefore, at a certain point, the text teacher stops providing sufficiently good supervision signals, introducing domain-specific noise that degrades the overall performance. Notice also how well the self-sustained method (\ourloss{} self) works with respect to the InfoNCE even without the initial help of the teacher, further proving that this self-distillation of scores helps stabilize the whole network. 

\subsubsection{Motion-to-Motion Retrieval Evaluation}

\begin{table}
\caption{\textbf{Motion-to-motion} retrieval using \textbf{joint-dataset} learning: training on KITML+HumanML3D, testing on HumanML3D}
\label{tab:resultsJDL-HumanML3Deval-m2m}
\begin{tabular}{llcc}
\toprule
\textbf{Protocol} & \textbf{Method} & mAP$\uparrow$ & nDCG$\uparrow$ \\
\toprule
\multirow[c]{4}{*}{\makecell[l]{(a) All}}
 & TMR~\cite{PBV23-TMR} & 0.728 & \bfseries 0.906 \\
 & MoT~\cite{MSFR23-sigir} & 0.704 & 0.899 \\
 & MoT++ & 0.725 & 0.903 \\
 & MoT++ self & \bfseries 0.734 & 0.905 \\
\cline{1-4}
\multirow[c]{4}{*}{\makecell[l]{(c) Dissim.\\subset}}
 & TMR~\cite{PBV23-TMR} & 0.850 & \textbf{0.922} \\
 & MoT~\cite{MSFR23-sigir} & 0.817 & 0.905 \\
 & MoT++ & \bfseries 0.853 & \bfseries 0.922 \\
 & MoT++ self & 0.845 & 0.918 \\
\cline{1-4}
\multirow[c]{4}{*}{\makecell[l]{(d) Small\\batches}}
 & TMR~\cite{PBV23-TMR} & 0.757 & 0.867 \\
 & MoT~\cite{MSFR23-sigir} & 0.723 & 0.850 \\
 & MoT++ & 0.763 & 0.863 \\
 & MoT++ self & \bfseries 0.777 & \bfseries 0.877 \\
\bottomrule
\multirow[c]{4}{*}{\makecell[l]{\bfseries average\\\bfseries over\\\bfseries protocols}}
 & TMR~\cite{PBV23-TMR} & 0.778 & 0.898 \\
 & MoT~\cite{MSFR23-sigir} & 0.748 & 0.885 \\
 & MoT++ & 0.780 & 0.896 \\
 & MoT++ self & \bfseries 0.785 & \bfseries 0.900 \\
\bottomrule
\end{tabular}
\end{table}

One of the potentially nice byproducts of the proposed \ourloss{} objective is the creation of a better uni-modal common space organization, which better places the motion features in the latent space.
This is a very desirable property, as it enables the same proposed \ourarchitecture motion encoder to be employed in the \textit{query-by-example} setup, which consists of effectively finding motions mostly resembling a given query motion example, as discussed in many previous works~\cite{SEBZ21-survey,JLHYJP23,LWFP21,SCA23-ecir}.

To quantitatively assess the ability of our model to perform motion-to-motion retrieval, we manually labeled the test set of the KITML dataset using a total of 90 different motion labels, which cluster the whole variety of samples present in KITML -- ranging from common classes \emph{walking} or \emph{jumping} to more particular motion classes like \emph{playingGuitar} or \emph{kneelingDown}. We then evaluated the motion-to-motion retrieval effectiveness employing classical binary relevance metrics, specifically mean Average Precision (mAP) \cite{zhu2004recall} and normalized Discounted Cumulative Gain (nDCG) \cite{jarvelin2002cumulated} with binary relevances.

The results are depicted in Table~\ref{tab:resultsJDL-HumanML3Deval-m2m}. Notice that we did not report the results for "All with threshold" protocol, as in this scenario, it would be particularly biased -- as all the motions having more than 0.95 textual similarity with the others also have the same label. We report both \ourarchitecture{} with the main hyper-parameter configuration (\ourloss{} 40-100), as well as the interesting case of fully self-supervised learning (\ourarchitecture{} self), where the teacher model $\mathcal{E}_t^\text{ref}$ is not employed. As you may notice, our method reaches the best results on almost all the protocols, achieving the best overall results on the averaged values. Notably, the best result in this scenario is often obtained using the ``\ourarchitecture{} self'' configuration. This may be attributed to the fact that the text model may introduce some text-specific biases during the optimization of the uni-modal manifolds, which act as slight noise in the motion domain. These results show that our model can effectively learn either a good cross-modal space, where text-to-motion and motion-to-text searches can be effectively performed, and a good uni-modal motion space suitable for the downstream query-by-example application scenarios.

\subsubsection{Visual Inspection of Text-to-Motion Retrieval Results}

\begin{figure}[t]
 \centering
 \begin{subfigure}[t]{\linewidth}
     \centering
     \includegraphics[width=.94\linewidth,page=2]{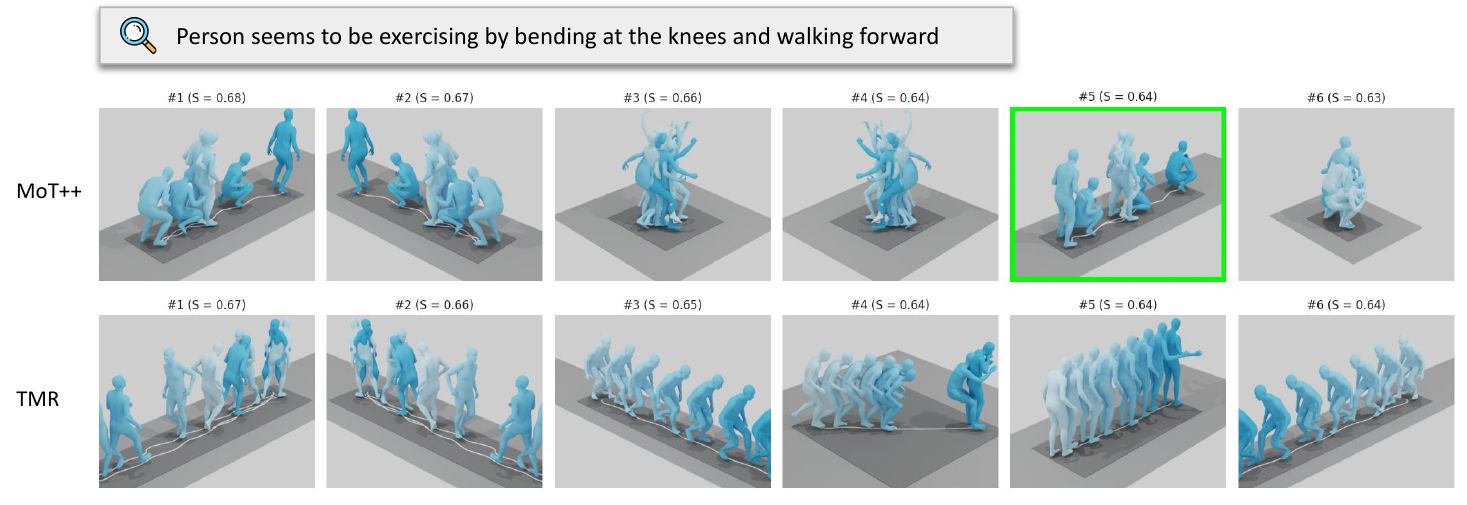}
     \caption{}
      \label{fig:t2m-examples-a}
 \end{subfigure}
 \begin{subfigure}[t]{\linewidth}
     \centering
     \includegraphics[width=.94\linewidth,page=3]{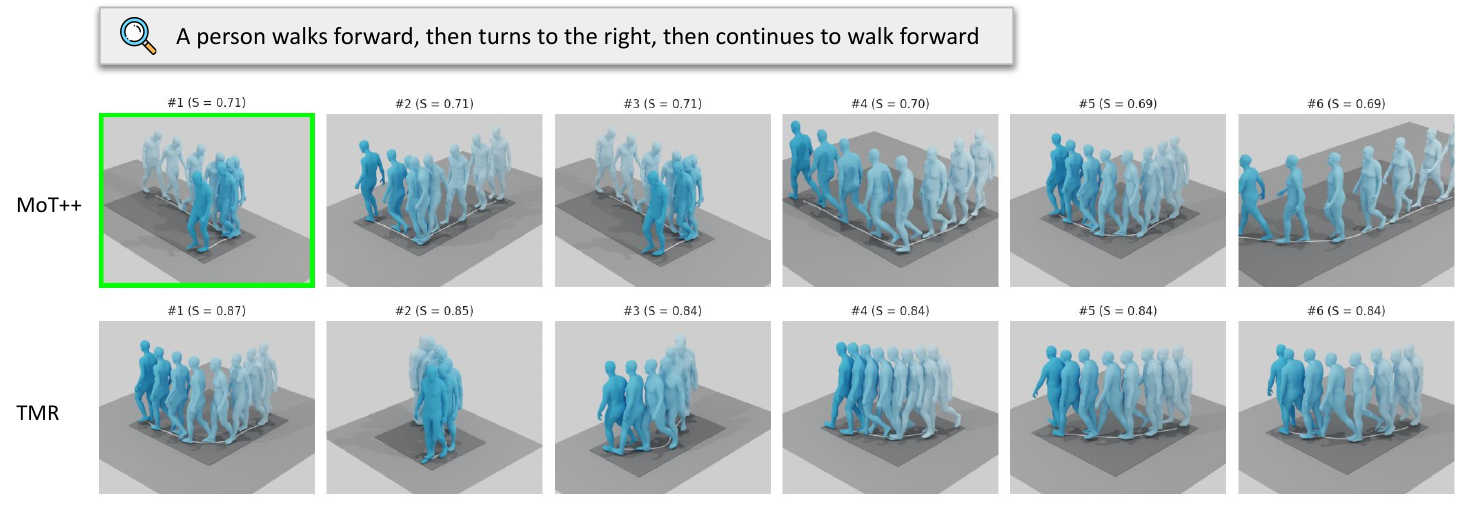}
     \caption{}
      \label{fig:t2m-examples-b}
 \end{subfigure}
 \begin{subfigure}[t]{\linewidth}
     \centering
     \includegraphics[width=.94\linewidth,page=4]{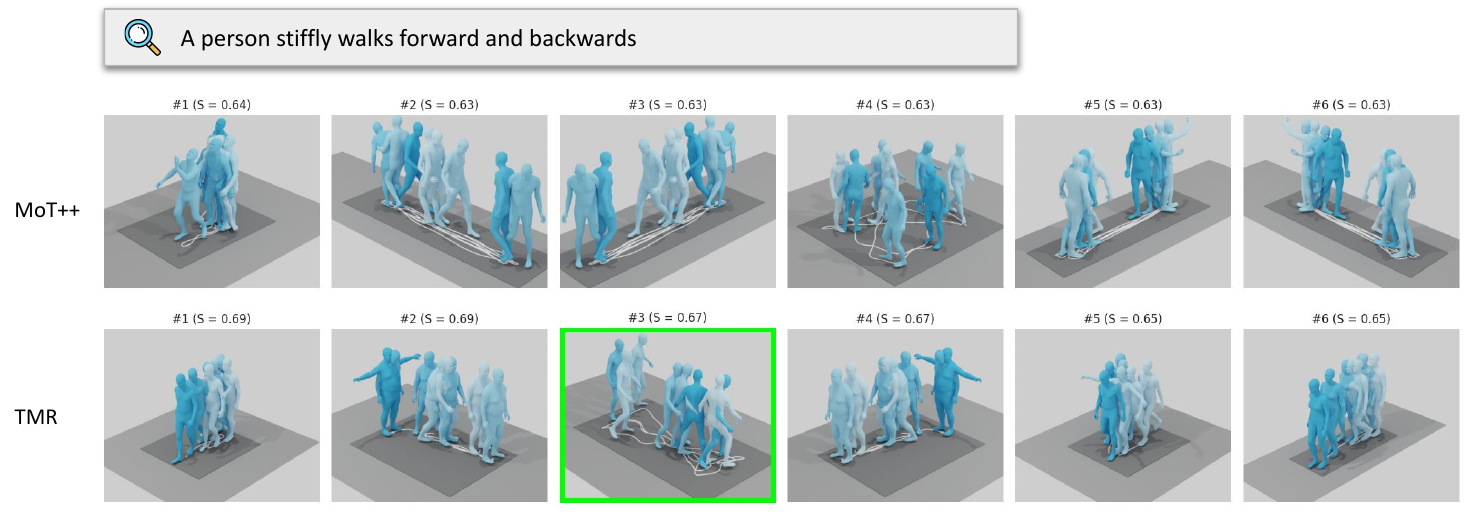}
     \caption{}
      \label{fig:t2m-examples-c}
 \end{subfigure}
 \caption{Qualitative examples on \textit{text-to-motion} retrieval. Samples (a), (b) show success cases in which \ourarchitecture{} can find the GT text better describing the motion within the first six results (highlighted using a green border), while TMR finds it at higher ranks. Sample (c), instead, shows a failure scenario in which \ourarchitecture{} cannot find the GT motion because it misses an important discriminative attribute (\textit{stiffly} in this case).}
 \label{fig:t2m-examples}
\end{figure}

\begin{figure}[t]
  \centering
  \includegraphics[width=.94\linewidth,page=6]{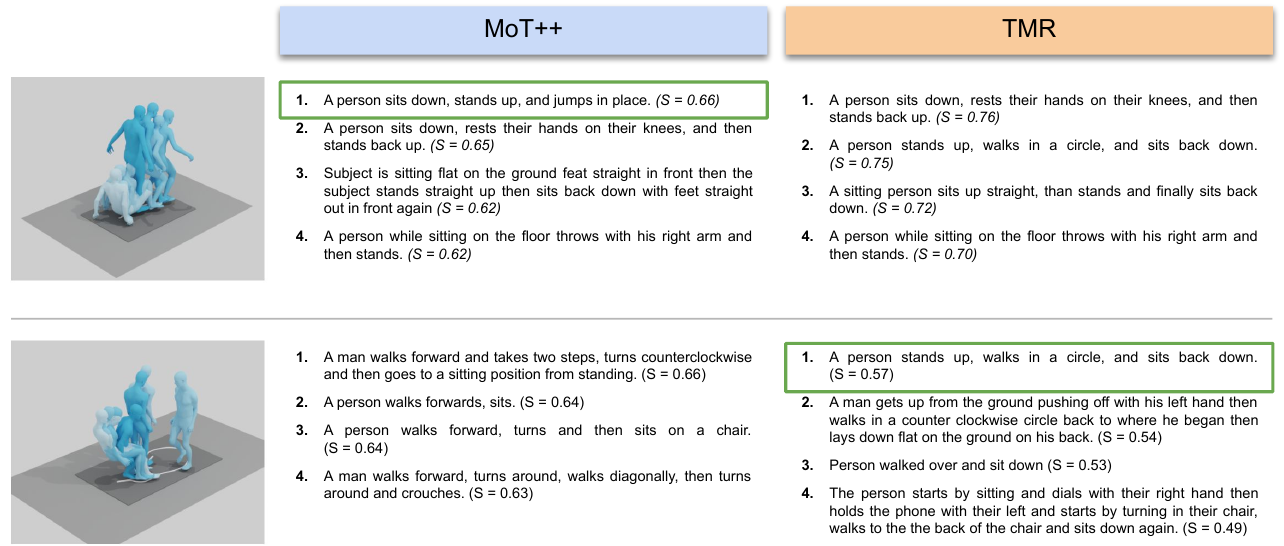}
  \caption{Qualitative results for \textit{motion-to-text} retrieval. The top example shows how \ourarchitecture\ can effectively retrieve the GT description in the first position. The second example, instead, shows a challenging case in which TMR is able to catch the correct text, although \ourarchitecture\ can still retrieve texts that are relevant to at least a subpart of the query motion.}
  \label{fig:m2t-examples}
\end{figure}

In Fig.~\ref{fig:t2m-examples}, we show some targeted success examples (Fig.~\ref{fig:t2m-examples-a} and Fig.~\ref{fig:t2m-examples-b}), as well as some failure case (Fig.~\ref{fig:t2m-examples-c}) of our method for text-to-motion retrieval, comparing \ourarchitecture{} with the state-of-the-art model TMR. Specifically, the first two subfigures show how our model can retrieve motions that better follow the provided fine-grained textual query. In particular, in Fig.~\ref{fig:t2m-examples-a}, our model better attends to the composite motion requested by the query (\textit{bending the knees} and \textit{walking forward}) in contrast to TMR which, instead, finds people just walking forward as the most relevant results. In Fig.~\ref{fig:t2m-examples-b}, we can assess the capability of \ourarchitecture{} to separate actions in the motion that happen in subsequent temporal order (first \textit{walk forward}, then \textit{turn right}, then \textit{walk forward} again). While this sequence is clear in \ourarchitecture, this is not always the case in TMR, which retrieves motions where the person simply performs a slight turn to the right. Instead, in Fig.~\ref{fig:t2m-examples-c}, we show a challenging example where our model, differently from TMR, cannot correctly understand fine-grained textual details, like \textit{stiffly}.

Similarly, in Fig.~\ref{fig:m2t-examples}, we report a success and a failure case for the motion-to-text retrieval scenario. Specifically, the top example shows how our method can retrieve the best-matching textual description as the first element, while TMR fails to capture the final action performed in the motion, which is \textit{jumping}.
Instead, in the lower example, we can notice how our model is missing the looping nature of the motion (\textit{the person is sitting, then stands up to walk in a circle, then sits back again}), despite anyways retrieving texts that correctly describe consecutive actions of the query motion (\textit{A person walks forwards, sits}).

%
%
%

\section{Conclusions}

In this work, we have incorporated cross-dataset and joint-dataset training for the text-to-motion retrieval task.
We have introduced the enhanced motion encoder \ourarchitecture{} and the new loss function \ourloss, which contributes to more effectively mitigating the gaps of datasets different in their nature.
We have demonstrated that our approach outperforms state-of-the-art approaches in the joint-dataset or cross-dataset scenarios. We have also demonstrated the generalization abilities and robustness of our approach, which achieves better results on the cross-dataset scenario (i.e., trained on HumanML3D and validated on KITML) in comparison with the narrowly-focused single-dataset scenario (i.e., trained and validated on KITML).
Future improvements include integrating additional modalities (e.g., video modality) and applying pair-wise text-to-modality learning to further boost the performance, similarly as in LanguageBind~\cite{ZLNYCWPJZLZLLY24}. 


\begin{acks}
This research was supported by the Ministry of the Interior of the CR project ``Automated digital data forensics lab for complex crime detection'' (No. VK01010147), by FAIR -- Future Artificial Intelligence Research - Spoke 1 (EU NextGenerationEU PNRR M4C2 PE00000013), by AI4Media -- A European Excellence Centre for Media, Society, and Democracy (EC, H2020 No. 951911), and by SUN -- Social and hUman ceNtered XR (EC, Horizon Europe No. 101092612).
\end{acks}

\bibliographystyle{ACM-Reference-Format}
\bibliography{references-Jan-disa,references-Jan,references-Nicola}




\end{document}